%% file: acl_latex.tex
\pdfoutput=1

\documentclass[11pt]{article}

\usepackage[preprint]{acl}

\usepackage{times}
\usepackage{latexsym}
\usepackage{xcolor}
\usepackage{float}
\usepackage{multirow}  

\usepackage[T1]{fontenc}

\usepackage[utf8]{inputenc}

\usepackage{microtype}

\usepackage{inconsolata}

\usepackage{graphicx}

\usepackage[most]{tcolorbox}
\usepackage{enumitem}
\usepackage{booktabs}
\usepackage{multirow}
\usepackage{tabularx}
\setlist[itemize]{noitemsep}

\usepackage{CJKutf8}
\usepackage[utf8]{inputenc}
\usepackage[vietnamese,english]{babel}
\usepackage[vietnamese=nohyphenation]{hyphsubst}
\newcommand{\vn}[1]{{\begin{otherlanguage}{vietnamese}#1\end{otherlanguage}}}
\newcommand{\zhs}[1]{{\begin{CJK}{UTF8}{gbsn}#1\end{CJK}}}
\newcommand{\zht}[1]{{\begin{CJK}{UTF8}{bsmi}#1\end{CJK}}}

\input{macros}

\title{Multilingual Retrieval Augmented Generation for Culturally-Sensitive Tasks: A Benchmark for Cross-lingual Robustness}

\author{Bryan Li, Fiona Luo\thanks{Equal second contribution, $^\dagger$Equal third contribution}, Samar Haider\footnotemark[1], Adwait Agashe\footnotemark[1], Tammy Li$^\dagger$, Runqi Liu$^\dagger$, \\ \textbf{Muqing Miao$^\dagger$, Shriya Ramakrishnan$^\dagger$,  Yuan Yuan$^\dagger$, Chris Callison-Burch} \\
	    University of Pennsylvania \\
        Philadelphia, PA, USA \\
	    {\tt \{bryanli, ccb\}@seas.upenn.edu } }

\begin{document}
\maketitle
\begin{abstract}
The paradigm of retrieval-augmented generated (RAG) helps mitigate hallucinations of large language models (LLMs). However, RAG also introduces biases contained within the retrieved documents. These biases can be amplified in scenarios which are multilingual and culturally-sensitive, such as territorial disputes. 
We thus introduce \bordirlines,\footnote{BorderLines for \underline{I}nformation \underline{R}etrieval and \underline{I}n \underline{R}eal \underline{L}ife} a dataset of territorial disputes paired with retrieved Wikipedia documents, across 49 languages. We evaluate the \textit{cross-lingual robustness} of this RAG setting by formalizing several modes for multilingual retrieval. 
Our experiments on several LLMs show that incorporating perspectives from diverse languages can in fact improve robustness; retrieving multilingual documents best \textit{improves response consistency} and \textit{decreases geopolitical bias} over RAG with purely in-language documents. 
We also consider how RAG responses utilize presented documents, finding a much wider variance in the linguistic distribution of response citations, when querying in low-resource languages.
Our further analyses investigate the various aspects of a cross-lingual RAG pipeline, from retrieval to document contents.
We release our benchmark and code to support continued research towards equitable information access across languages.\footnote{Links for the dataset \href{https://huggingface.co/datasets/borderlines/bordirlines}{here}, and for the code \href{https://github.com/manestay/bordIRlines}{here}.}

\end{abstract}

\input{sections/01_intro}
\input{sections/02_background}

\input{sections/04_dataset}
\input{sections/05_setup}

\input{sections/06_results}
\input{sections/07_more_exp}
\input{sections/08_related}

\input{sections/09_conclusion}

\bibliography{custom}

\input{sections/appendix}

\end{document}

%% file: macros.tex
\newcommand{\borderlines}{\textsc{BorderLines}}
\newcommand{\bordirlines}{\textsc{\textcolor{RoyalBlue}{Bord}\textcolor{OliveGreen}{IRL}\textcolor{RoyalBlue}{ines}}}
\newcommand{\noir}{\texttt{no\_ir}}
\newcommand{\qlang}{\texttt{qlang}}
\newcommand{\qlangen}{\texttt{qlang+en}}
\newcommand{\enonly}{\texttt{en\_only}}
\newcommand{\en}{\texttt{en}}
\newcommand{\rellangs}{\texttt{rel\_langs}}
\newcommand{\swapdocs}{\texttt{swap\_docs}}
\newcommand{\langen}{\texttt{en}}
\newcommand{\control}{\texttt{control}}

%% file: sections/01_intro.tex
\section{Introduction}

Large language models (LLMs) have gained widespread adoption due to their remarkable text generation capabilities \cite{gpt-4, llama3, commandr}. Despite this, their tendencies to hallucinate and their static knowledge from training-time impacts their applicability for knowledge-intensive tasks. The paradigm of retrieval-augmented generation (RAG) addresses these pitfalls by using an information retrieval (IR) system to access documents from an external database, enabling  an LLM to generate grounded and more up-to-date responses \cite{rag}. RAG can be particularly useful in multilingual settings by helping an LLM generate more accurate responses ~\cite{asai2022mia}, and by allowing it access to different perspectives on more subjective, nuanced issues~\cite{li2024uncovering}.

While several recent studies have investigated RAG in small-scale multilingual settings~\cite{sharma2024faux,wu2024not}, they only consider high-resource languages and do so in artificially construed scenarios with synthetic documents. A wider approach has been taken in work on open-retrieval multilingual QA~\cite{tydiqa,cora}, but these primarily focus on simple fact-seeking questions where right answers are easily memorized by LLMs, in which cases, RAG can even be harmful~\cite{mallen2023not}. 

In this work, we investigate the impact of RAG in a massively multilingual setting. We focus on queries about \textit{territorial disputes}, which exist at the intersection of cultural and linguistic fault lines. Although these queries are fact-seeking in nature, their answers are highly dependent on one's perspective.
\citet{li2024land} found that LLMs' responses are inconsistent depending on the language of interaction, and that such inconsistency increases the societal risks of LLMs by promoting divisive viewpoints across cultures.
Meanwhile, a RAG system's answers are highly dependent on the \textit{selection of documents}, particularly over different perspectives in the multilingual setting.
Many questions arise: how does the linguistic composition of the set of documents impact responses? Does sourcing information from different languages increase or decrease consistency? And is presenting conflicting information to LLM's base preferences better expressed in certain languages?

\begin{figure*}[t!]
    \centering
    \includegraphics[width=0.98\linewidth]{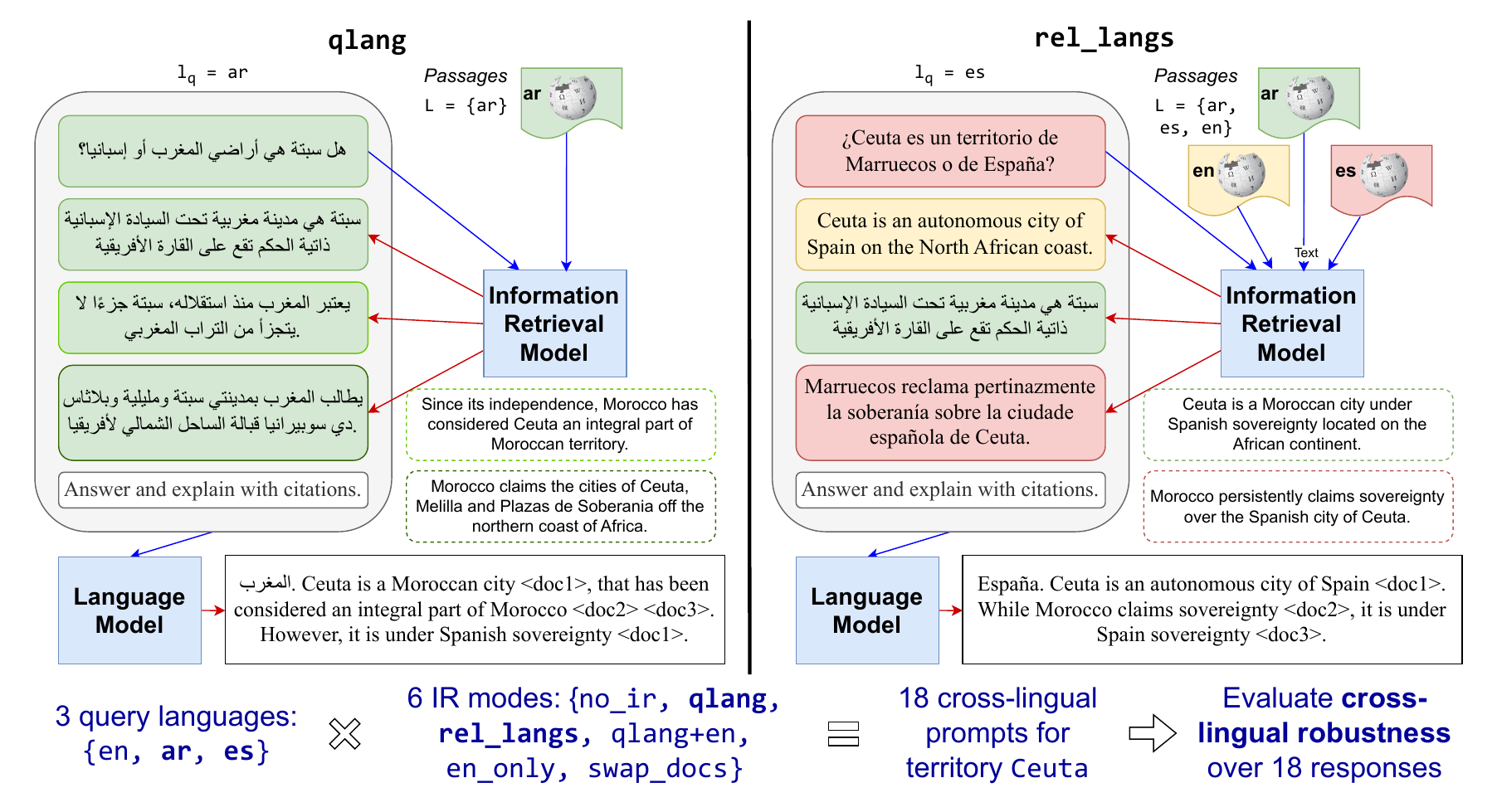}
    \caption{Illustration of 2 cross-lingual RAG prompts from the \bordirlines\ dataset, showing differences in a) retrieved documents by the IR model and in b) responses by the LLM. For a territory such as ``Ceuta'', we create several prompts across languages and IR modes; responses to them are compared to responses from the \noir\ setting of \borderlines~\cite{li2024land}. Evaluation of \textit{cross-lingual robustness} is over the set of responses.}
    \label{fig:xlrag_2modes}
\end{figure*}

Our study makes the following contributions:
\begin{itemize}
    \item We design the \bordirlines\ dataset, which pairs queries from an existing territorial dispute dataset with 7.4k retrieved Wikipedia passages across 49 languages.
    \item We define five \textit{IR modes} for retrieving multilingual sets of documents, each of which reflects real-world information access needs.
    \item We use \bordirlines\ to systematically evaluate the \textit{cross-lingual robustness} of RAG with various LLMs. Most notably, retrieving over multilingual documents \textit{improves cross-lingual consistency} and \textit{decreases geopolitical bias} vs. over in-language documents.
    \item We find that for LLM responses' citations of documents, low-resource languages demonstrate much wider variability in citation rates than high-resource languages.
    \item We perform further analyses on all aspects of cross-lingual RAG. On IR, we find a preference toward retrieving query-language documents; on document contents, we find LLM responses can selectively interpret the same documents to fit their own viewpoints.
\end{itemize}


 

%% file: sections/02_background.tex
\section{Task Definitions}

\subsection{Cross-lingual RAG}
We construct a RAG pipeline consisting of a \textit{retrieval} stage using an IR system and a \textit{generation} stage using an LLM. Both use off-the-shelf multilingual models, without further fine-tuning. The cross-linguality in the setting is twofold: in the multilingual queries, and in their pairing with documents retrieved from a multilingual database.


\label{sec:task_form}
\paragraph{Territorial Disputes Task} We adapt the task formalization from \borderlines~\cite{li2024land}. We consider a territory $t$ that has a set of claimant countries $C=\{c_1, c_2, ...c_k\}$. For each $t$, \borderlines\ has a multilingual query set (MQS) which consists of multiple-choice questions in each of the claimant's languages: $\text{MQS}=\{q_{l_1}, q_{l_2}...q_{l_k}\}$. 

\paragraph{Retrieval} Given a query $q$, we use an information retrieval (IR) system to retrieve the top-k most relevant documents $D=\{d_1, d_2, ..., d_k\}$ from an external database $KB$. $KB$ is multilingual, being a concatenation of databases in different languages, depending on the IR mode (described ahead).

\paragraph{Generation} A query $q$ and relevant documents $D$ are composed into a prompt $P$. The LLM's task is to generate an answer $a$ for $P$, informed by both its parametric memory and the external retrieved information.\footnote{As $q$ is in language $l$, the generated $a$ should also be in $l$.} 
A multilingual response set (MRS) consists of all responses for a territory $t$; we evaluate cross-lingual robustness over an MRS.

\subsection{Information Retrieval (IR) Modes}
\label{sec:ir_mode}
We define five IR modes, each specifying the languages of documents to retrieve, given a query in $l_q$. These are formalized in Table~\ref{tab:ir_modes}. Each IR mode is motivated by some real-world  information access needs a user may have.
\textbf{\qlang} retrieves documents in the query (user's) language.
\textbf{\rellangs} retrieves documents in the user's language, English and any other relevant language.\footnote{For example, the \rellangs\ mode for Ceuta would have $KB=KB_\langen \cup KB_{\text{ar}} \cup KB_{\text{es}}$.}
\textbf{\qlangen} retrieves information in the query language and English.
\textbf{\enonly} retrieves only English.
Finally, \textbf{\swapdocs} adversarially selects documents a language besides the query language, as a test of whether information from differs cross-lingually.

\begin{table}[t!]
\centering
\small
\begin{tabularx}{\linewidth}{rX}
    \toprule
    \textbf{IR Mode}& \textbf{Retrieval Languages $L$} \\ \midrule
    \qlang     & $[l_q]$             \\
      $\hookrightarrow$ & monolingual IR in query lang \\
    \rellangs & $[l_q, l_\langen] + [L_\text{relevant}]$ \\
    $\hookrightarrow$ & multilingual IR in all \textit{relevant languages} \\ \midrule
    \qlangen  & $[l_q, l_\langen]$         \\
    $\hookrightarrow$ & multilingual IR in query lang and \langen \\
    \enonly         & $[l_\langen]$   \\$\hookrightarrow$ & cross-lingual IR in \langen \\
    \swapdocs         & $[l_{\text{not\_q}}]$   \\$\hookrightarrow$ & cross-lingual IR in  non-query lang \\
    \bottomrule
\end{tabularx}
\caption{For each IR mode, this depicts the retrieval languages $L$ for a query $q$ in language $l_q$. $\hookrightarrow$ indicates the descriptions. $L$ is a set where each $l$ is unique. We mainly compare the first 2 modes, and later the other 3.}
\label{tab:ir_modes}
\end{table}


\subsection{Cross-lingual Robustness}
\label{sec:xlrobustness}

\paragraph{Attributes}
\citet{li2024land} defines cross-lingual robustness as the ability of a language model to reliably produce the same answer to an underlying fact-seeking query, regardless of the prompt phrasing or language. We adopt this definition and extend it to the RAG setting by considering how answers are affected by multilingually-sourced documents in the prompt. The attributes of robustness considered by this work are as follows:

\begin{itemize}
    \item \textit{Factuality} is how well an LLM's responses align with a knowledge base's ground truth. In our work, we only measure factuality in English, the most well-represented language.
    
    \item \textit{Consistency} measures how consistent an LLM's responses are when asked the same query in different languages.
    In other words, consistency measures how often answers vary depending on the query language.
    
    \item \textit{Geopolitical Bias} is the tendency to report geopolitical knowledge differently depending on the language of interaction. 
    For example, geopolitical bias occurs when, for some territory, querying in Spanish returns ``Spain'' (or ``Espa\~{n}a''), while in English returns ``UK''.
\end{itemize}



We note that \textit{geopolitical bias} is specifically defined for the territorial disputes task. While related to consistency, which only considers whether responses vary across languages, geopolitical bias is more specific -- it considers whether language influences territory claimant judgments. \textit{Factuality} and \textit{consistency} are more general and can be applied to other cross-lingual tasks.


\paragraph{Comparison to Factual Robustness}
Prior work studying robustness in RAG has focused on \textit{factual robustness},
The two attributes are hallucination rate (generating non-existent information) and error rate (failing to use relevant information), construed as binary classification tasks~\cite{adlakha2024robustness, chen2024robustness,nomiracl}.

Although related, our work evaluates differing aspects of robustness.\footnote{Appendix~\ref{sec:why_robustness} provides the rationale for why our cross-lingual attributes are of robustness.} Factual robustness
considers \textit{a single interaction}, and how a response uses provided documents. Cross-lingual robustness considers \textit{multiple interactions}, and how responses differ given different queries and languages. Cross-lingual robustness can hence be applied to LLM's responses with or without RAG.


%% file: sections/04_dataset.tex
\section{The \bordirlines\ Benchmark}
\begin{figure}[t!]
    \centering
\includegraphics[width=.98\linewidth]{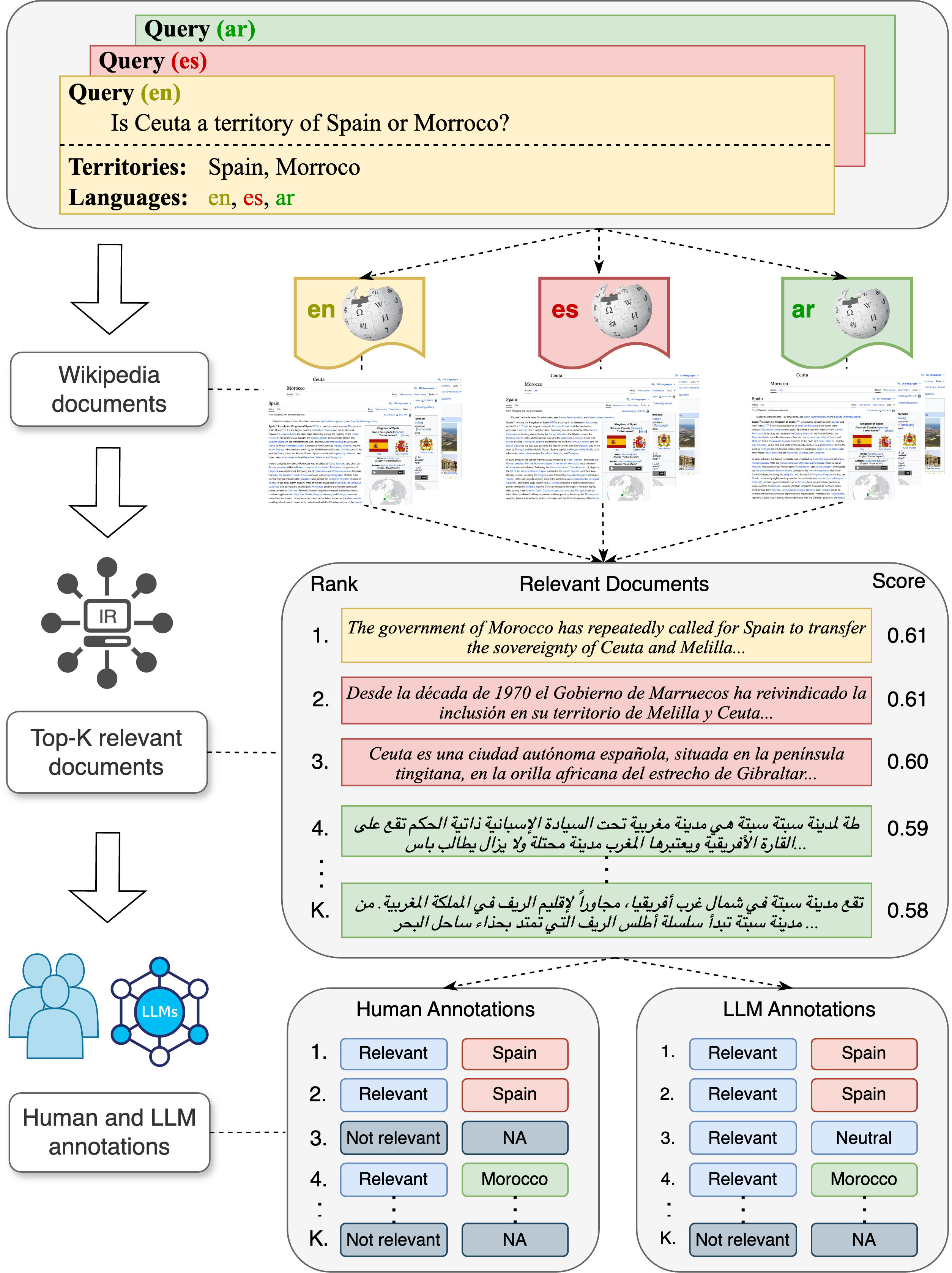}
    \caption{Illustration of data collection for a \bordirlines\ entry on the territory ``Ceuta''.  Depicted is the \rellangs\ IR mode, which considers Wikipedia documents in $L = [\text{en, ar, es}]$. For each query (queries are sourced from \borderlines), we use IR to obtain the top-10 most relevant documents. Each document is then annotated for relevance and territorial viewpoint.
    }
    \label{fig:wiki_collection}
\end{figure}


\subsection{Dataset Description}
\bordirlines\ is a multilingual dataset covering 49 languages. It builds on top of the \borderlines\ dataset~\cite{li2024land}, which consists of 720 queries for 251 disputed territories. \bordirlines\ extends the cross-lingual RAG setting by associating each query with several sets of retrieved Wikipedia documents (1 set per IR mode). Each set consists of the 10 most-relevant documents, as scored by IR systems. Each query-document pair is also annotated for its \textit{relevance} and \textit{viewpoint}.

Figure~\ref{fig:wiki_collection} illustrates the dataset construction and annotation. 
Appendix Table~\ref{tab:article_counts} provides the detailed, per-language dataset statistics. In total, there are 19,916 unique query-document pairs, drawn from 7,436 passages from 905 Wikipedia articles.

\subsection{Dataset Construction}
\label{sec:construction}
\paragraph{Source: Wikipedia}
We source the documents from Wikipedia, an online encyclopedia available in multiple languages. We do this for two reasons. First, Wikipedia is the primary source used by prior benchmarks on retrieval and QA~\cite{xorqa,miracl}. Second, Wikipedia articles go through
a set of community-enforced standards, such as neutral POV. Therefore, we expect articles in different languages to report from similarly ``neutral'' viewpoints, and minimize culturally specific biases. Still, as articles are independently authored, this is not guaranteed. This motivates us to annotate each document's content as described ahead.

\paragraph{Information Retrieval Setup} We segmented articles into paragraphs by splitting articles on double newlines, and used these paragraphs as the documents.\footnote{We thus use the terms `passage' and `document' interchangeably, following common usage with IR over Wikipedia.} In lieu of retrieving over the entirety of the Wikipedia $KB$, we index only the articles most likely to find relevant information -- on the territory, and on each claimant (from the annotations of \borderlines). This greatly reduces the search space and the likelihood of retrieving non-relevant documents. As described in~\ref{sec:task_form}, we perform multilingual IR, using the query $q$ to retrieve over a $KB$ whose linguistic composition depends on the IR mode.

\subsection{Annotating Documents for Content}
\label{sec:content}
To improve the \bordirlines\ benchmark's usability for IR, 
 each query-document pair is annotated for two content factors: \textit{relevance} -- whether it directly discusses the status of a territory \footnote{Note that this is a very strict definition of relevance.} -- and \textit{territorial viewpoint}  -- which claimant it supports (or Neutral or N/A). These were obtained in two stages. First, we performed a small-scale human annotation effort, on 5 languages: English, Chinese, Hindi, Spanish, and Korean\footnote{We chose these 5 languages due to annotator availability, as well as their diversity in language families and scripts.}. Second, informed by the human effort, we designed an LLM-as-annotator approach to annotate the full dataset. We describe each annotation stage in brief below, and provide the full details in Appendix~\ref{sec:annotation_details}.

\paragraph{Human Annotation}
For the human annotation effort, we recruited from a graduate-level computer science course annotators who are bilingually fluent in English and one of the other 5 languages. For each task, an annotator is shown a query, as well as 10 retrieved documents from a given IR system and mode. For each document, annotators are asked to fill in two multiple-choice questions, on the relevance and on the viewpoint (so each task is 20 annotations). We assigned three annotators per task, and use the majority label for each annotation.

We performed several iterations, informed by participant feedback, before finalizing instructions. Annotations were requested for 620 pairs; after accounting for overlapped retrievals and quality control, 543 pairs were annotated in total.

 \paragraph{LLM Annotation}
To scale up annotation to the full \bordirlines\ dataset of 19k pairs in 49 languages, we designed a two-stage LLM-as-annotator pipeline. We use GPT-4o, and use exemplars drawn from the human-annotated subset for few-shot prompting.
First, we prompt for relevance. Then, for query-document pairs marked as relevant, we prompted separately for viewpoint. On relevance, the LLM judged 21.5\% of retrievals to be relevant.\footnote{Low relevance is expected with IR over real-world documents. We emphasize that the results of RAG do not depend at all on these annotations, which are a meta-evaluation of the quality of multilingual IR systems for this task.} Agreement between LLM-annotated and human-annotated labels  was high, at 76\% F1 agreement rate. On viewpoint, however,  there was a low agreement rate.\footnote{Our manual analyses in Appendix~\ref{sec:annotation_f1_details} revealed the subjectivity of the viewpoint annotation task (for both LLM and human), given the subtle and nuanced texts.}

%% file: sections/05_setup.tex
\section{Experimental Setup}

\paragraph{IR Systems Used}
We consider two multilingual IR systems. We use \textbf{OpenAI embeddings}~\cite{openai2024embedding} to embed documents and queries,\footnote{\texttt{text-embedding-3-large}} and use cosine similarity as the distance metric.
We also use the open-source \textbf{M3-Embedding}, following the official implementation of the hybrid retrieval system~\cite{bgme3}.
Both systems had reasonable performance for multilingual IR. We mainly report results using OpenAI embeddings as its documents achieved higher downstream performance than M3.

\paragraph{LLMs Used} We consider several representative LLM families with multiple model sizes, bolding the ones studied in the main text. For closed-source, we use \textbf{GPT-4o} and \textbf{GPT-4o-mini}~\cite{gpt4}. For open-source, we use Llama 3 \{1B, 3B, \textbf{8B}\}~\cite{llama3}, as well as Command-R \{7B, \textbf{35B}\}~\cite{commandr}.

\subsection{Evaluation Methodology}
\label{sec:eval_method}

\paragraph{Response Formats}
We use 2 different prompt templates to elicit responses in two formats.\footnote{Refer to Appendix~\ref{sec:full_prompts} for the full prompts used.} For the \textit{direct} format, the instructions are short, and specify to output one of the multiple-choice selections.
For the \textit{citation} format, the instructions are more detailed, and specify a multipart output: a selection, and n natural-language explanation which cites documents IDs (e.g., \texttt{<doc1>}).
For each setting, we perform 10 runs with 10 fixed random seeds and temperature=0.5, and report the average scores and 95\% confidence intervals.

\paragraph{Metrics for Cross-lingual Robustness}
We adopt the cross-lingual robustness evaluation suite from \citet{li2024land}. This is based on the concurrence score metric (CS), which is a binary accuracy between two strings -- 1 if equal, 0 otherwise.  
The derived metrics are:\footnote{Refer to  Appendix~\ref{sec:cs_metrics} for the metrics' formulas and details.}
\begin{itemize}
    \item \textbf{Knowledge-base CS} $\uparrow$: compares the claimants from the KB and the English response. This is our measure of \textit{factuality}.
    \item \textbf{Consistency CS} $\uparrow$: compares all pairwise responses in an MRS. This is our measure of cross-lingual \textit{consistency}.
    \item \textbf{$\Delta$ CS} $\downarrow$: the difference of the controller language preference and the non-controller ones. This is our measure of \textit{geopolitical bias}.
\end{itemize}

Each CS metric is calculated over the multilingual response set (MRS) for a territory. For the 720 queries, there are 251 territories, and thus 251 MRSs. We will then compare the averaged score across all territories, for each RAG setting consisting of an IR mode and an LLM. 

\paragraph{Metrics for Citation Analysis}
We calculate two metrics to analyze responses in the citation format.
Considering a prompt with 10 documents, the \textit{inclusion rate} ($nr$) for a language $l$ is the fraction of documents in that language. Considering the citation format response, The \textit{citation rate} ($cr$) of $l$ as the rate in which its documents are cited. 

We assign a $cr$ and $nr$ to a language by averaging over all its queries. We say there is a query-language preference when $cr >> nr$, and an anti-preference when $cr << nr$.






%% file: sections/06_results.tex
\section{Experimental Results}
We report results on our two main experiments: on cross-lingual robustness for the direct response format, and on LLM's use of documents in the citation response format. We consider 4 LLMs here, and all 7 in Appendix Table~\ref{tab:full_results}.

\begin{figure}[t!]
    \centering
    \includegraphics[width=0.98\linewidth]{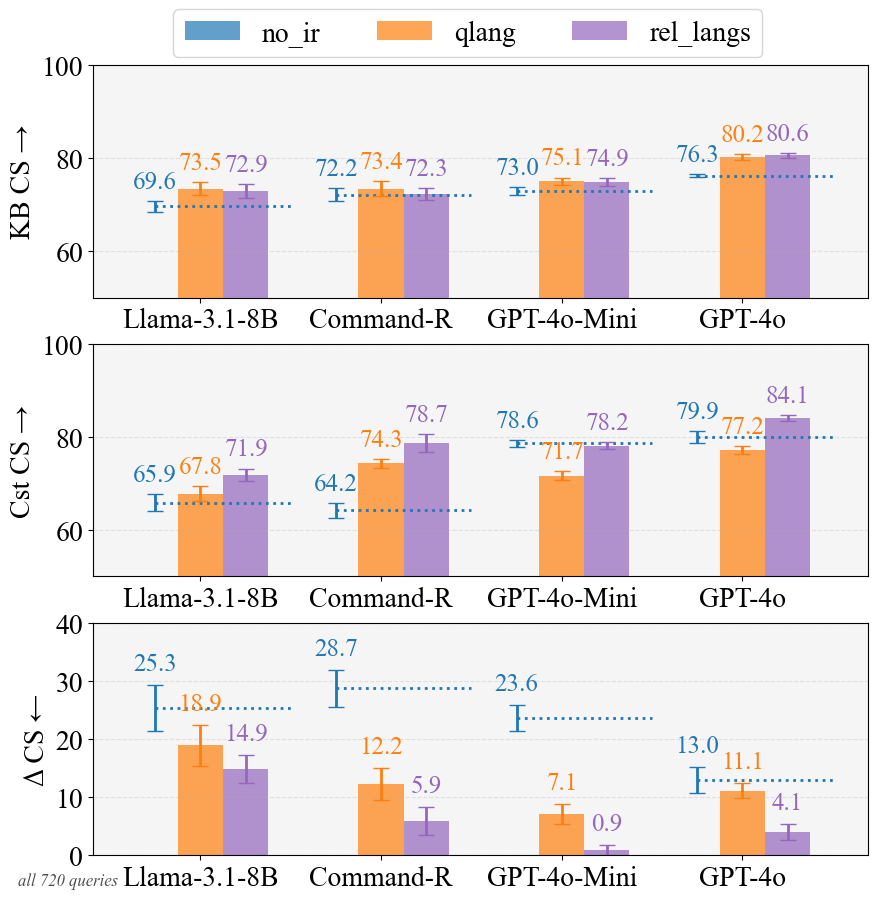}
    \caption{Results for the concurrence score (CS) metrics, which measure attributes of cross-lingual robustness: KB CS for \textit{factuality}, Cst CS for \textit{consistency}, and $\Delta$ CS for \textit{geopolitical bias}. Within each subplot, we display the results for 3 IR modes: \qlang\ and \rellangs\ with bars, and the \noir\ baseline with the dotted line. The score for each bar is the average over 10 runs, and is shown with its 95\% confidence interval.} 
    \label{fig:CS_metrics}
\end{figure}

\subsection{Results on Cross-lingual Robustness}
\label{sec:direct_eval}

Figure~\ref{fig:CS_metrics} presents the results for each model and IR mode combination, for each CS metric. 
\noir\ is the baseline without IR and using only the LLM's parametric knowledge. \qlang\ is RAG with in-language IR, and \rellangs\ is RAG setting with multilingual IR (MLIR).
Recall that KB CS is calculated over English queries and measures factuality. 
Cst CS and $\Delta$ CS both are calculated over all language queries, and measure different but related attributes of consistency: general (consistency) and systematic (geopolitical bias) respectively.


\paragraph{Factuality generally increases with RAG}
The top panel concerns only English queries (but multilingual documents). Compared to \noir\, both IR modes reliably boost KB CS scores for all LLMs; this concurs with the general view of RAG improving factuality of responses. \qlang\ outperforms \rellangs\ for 3 of the 4 LLMs, though the intervals overlap in all cases (we can see that intervals for KB CS are all narrow).
We also see that the the largest gains are for the least and most factual LLMs, Llama and GPT-4o.


\paragraph{Consistency varies with RAG}
The middle panel considers multilingual queries and responses. \qlang\ has mixed effects -- Cst CS greatly increases for  for Command-R ($64.2 \rightarrow 74.3$), while decreasing for GPT-4o-mini ($78.6\rightarrow 71.7$) and GPT-4o ($79.9 \rightarrow 77.2$).
\rellangs\, in contrast, has a positive effect, especially for the least consistent model, Command-R ($64.2 \rightarrow 78.7$). These findings underscore the importance of our study on the linguistic \textit{selection of documents} for RAG. By retrieving over multiple languages, the set of LLM's responses across a query posed in different languages can be, in fact, be more consistent.
 

\paragraph{Geopolitical bias reliably decreases with RAG}
The bottom panel also considers multilingual queries and responses, but focusing on the geopolitical bias aspect. We see that both RAG settings reduce $\Delta$ CS and thus geopolitical bias. 
$\rellangs$\ is more effective than \qlang, which again shows that having the multilingually-sourced perspectives is helpful in reducing LLM inconsistencies. We see that the largest decreases with Command-R ($28.7 \rightarrow 12.2, 5.9$), and GPT-4o-mini ($23.6 \rightarrow 71.9, 0.9$; observe the neutralization of bias with \rellangs\ here). Furthermore, we see that $\Delta$ CS in the \noir\ modes have wide confidence intervals, due to the cultural-sensitivities of the task; but with RAG, they are greatly narrowed.

\paragraph{LLMs display different sensitivities to RAG} We also observe that in terms of RAG affecting responses, Llama sees the smallest changes, and Command-R is the most affected. This aligns with the fact that Command-R was explicitly trained for RAG, while Llama was not. The greater sensitivity invites more caution into selection of documents -- selecting especially biased (or factually incorrect) documents will harm Command-R's robustness more than other models, and conversely for less biased documents.


\paragraph{Discussion}
Overall, utilizing RAG improves cross-lingual robustness for all three attributes over the \noir\ baseline. On one hand, this is expected, given the purpose of RAG in grounding responses to the information provided in documents. Here, the documents and the original task are both sourced from Wikipedia, as with prior work.

The use of CS metrics enable quantified insights into how different attributes of robustness are influenced by the IR mode, and the LLM used.
The most interesting finding is in the success of \rellangs. Prior works studied \textit{synthetic} questions and documents, and raised the potential challenges RAG systems face in reconciling multilingual knowledge conflicts. But our investigation shows that, in a \textit{real-world setting}, concerns for such conflicts may be over-stated, at least given a more reliable source as Wikipedia. Multilingual RAG overall increases cross-lingual consistency of responses, while the in-language RAG (\qlang) setting that lowers this robustness attribute.


\begin{figure}[t!]
    \centering
    \includegraphics[width=0.98\linewidth]{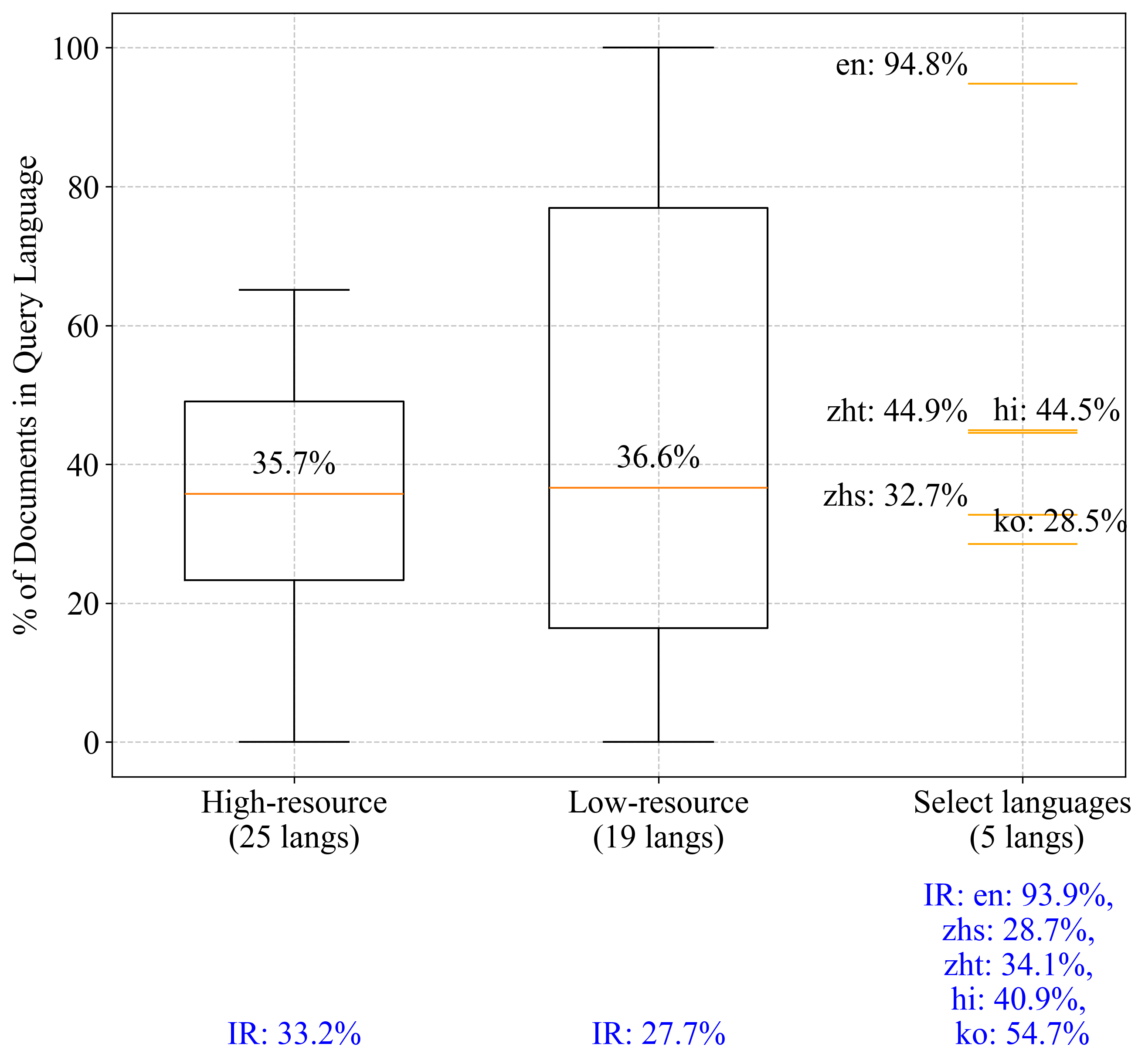}
    \caption{The $l_q$ RAG citation rates (\textcolor{orange}{orange}) and inclusion rate (\textcolor{blue}{blue}), for each language in the \rellangs\ setting, using GPT-4o. Results are grouped into high-resource and low-resource, and selected languages.}
    \label{fig:citation_boxplot}
\end{figure}

\subsection{Results on RAG Citations}
\label{sec:citation_eval}
Using responses from the citation format, we investigate how an LLM provided with documents uses them in its response, and whether it prefers to cite certain languages' documents  over others. 
We thus consider the \rellangs\ IR mode.
We group the 49 languages into 3 groups: low-resource (LR), high-resource (HR), and selected languages (the highest resource languages, and the ones we had performed some human annotations on). 

Results for \rellangs\ and GPT-4o are shown in Figure~\ref{fig:citation_boxplot}. 
Considering the box-plots, we see similar median citation rates for HRs and LRs ($cr \approx 36\%$). LRs have an overall in-language preference ($nr=27.7\%$), while HRs do not ($nr=33.2\%$). 
LR also has a much wider spread, from 0-100\%, while HR has a smaller spread from 20-60\%.

On the selected languages, the most striking finding is that \rellangs\ with an English query is barely multilingual, with $cr=94.8\%, nr=93.9\%$. For ko, there is a strong anti-preference ($cr=28.5\%, nr=54.7\%$), while 
for zht there is a preference ($cr=44.9\%, nr=34.1\%$).

\paragraph{Discussion} 
While prior studies on HR languages found LLMs prefer to use query-language and English documents, we found more nuances exist when considering 49 languages.
Low-resource languages are susceptible to large swings in (anti-)preference. 
We also note the self-fulfilling nature of high $nr$ enabling high $cr$. Our study of LLM's citation preferences emphasize the need for more inclusive retrieval techniques and generative models for underrepresented languages.

%% file: sections/07_more_exp.tex
\section{Further Analysis}
\subsection{Linguistic Biases of Cross-lingual IR}
\label{sec:ir_bias}

\begin{figure}[t!]
    \centering
    \includegraphics[width=\linewidth]{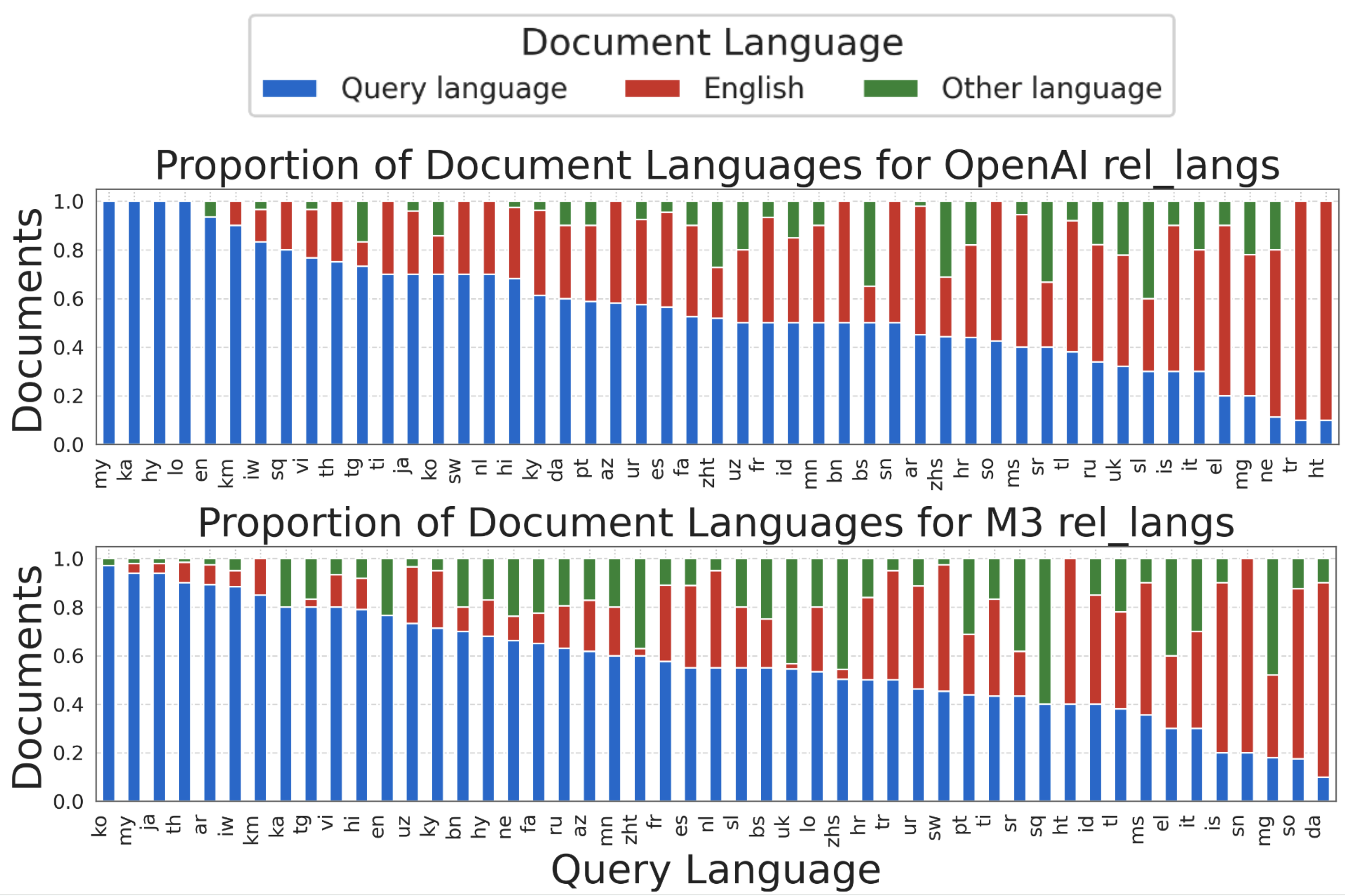}
    \caption{Proportion of document languages retrieved per query language, using OpenAI and M3 embeddings.}
    \label{fig:openai_m3_docs}
\end{figure}

\paragraph{IR systems prefer in-language documents} 
For OpenAI in the \rellangs\ setting, a query language document is $1.29$ times more likely to be retrieved than a non-native document, increasing to $1.64$ times for M3. However, this varies greatly across languages, shown in Figures \ref{fig:openai_ir}, \ref{fig:m3_ir}. This finding, that IR systems have a preference towards in-language retrieval bias, is corroborated by \citet{sharma2024faux, wu2024not}.

\paragraph{OpenAI embeddings prefer English documents}
Figure \ref{fig:openai_m3_docs} shows a per-language breakdown of retrieved document languages for the two IR systems, OpenAI and M3. On average, OpenAI retrieved 1.72 times more English documents than M3 (1885 for OpenAI vs. 1102 for M3). We hypothesize that this could be due to the training of M3 having a larger focus on multilinguality. However, as M3 embeddings scored worse on the CS metrics, this highlights the challenges of multilingual RAG for current LLMs.


\subsection{Other Cross-lingual Robustness Results}
We summarize the findings of these experiments and give further details in Appendix~\ref{sec:addtl_ir_mode} and \ref{sec:addtl_rel}.
\paragraph{Additional IR Modes} 
Appendix Figure~\ref{fig:CS_metrics_all} shows results for all 6 IR modes. We consider the 3 other IR modes. \qlangen\ is similar to \qlang, and \enonly\ intuitively has the highest overall consistency.
The most interesting is with \swapdocs, where given a query in $l_1$, we retrieve documents from $l_2$, This reverses geopolitical bias, such that $\Delta$ CS is \textit{negative}, showing that the LLMs can be strongly influenced by the cross-lingual perspective differences of Wikipedia articles.

\begin{figure}[t!]
    \centering
    \includegraphics[width=0.98\linewidth]{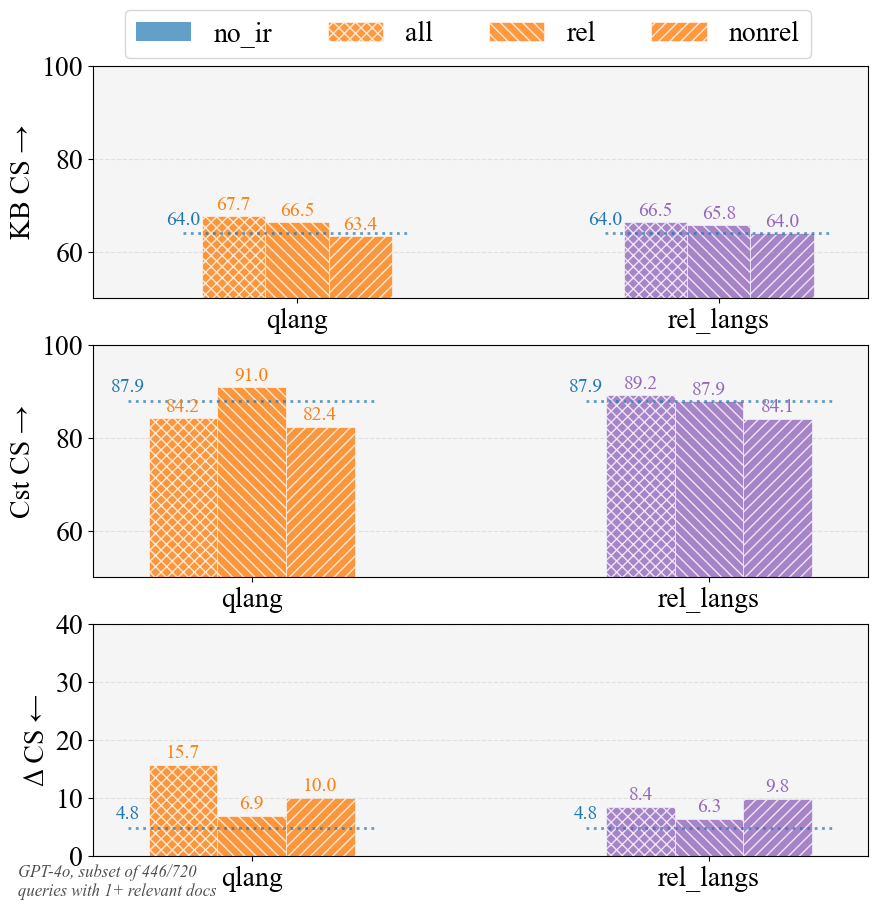}
    \caption{CS metrics for \qlang\ and \rellangs\ modes with GPT-4o.}
    \label{fig:relevance}
\end{figure}

\paragraph{Controlling for Relevance}
We utilize the annotations on relevance to compare: \texttt{all} 10 documents, \texttt{rel}evant only or \texttt{nonrel}evant only. We use GPT-4o and \qlang\ and \rellangs\ modes, and results are shown in Figure~\ref{fig:relevance}.
\texttt{all} has slightly higher factuality, suggesting that due to LLMs' background knowledge on disputes, they can use the helpful parts and ignore the non-helpful parts. Consistency is similar between all settings. Geopolitical bias for \qlang\ is much lower for \texttt{rel} and \texttt{nonrel}, which suggests that the classes of documents are in sum providing more biased opinions. These unpredictable effects of relevance may speak to the factual robustness explored in prior work, and that RAG may be causing some tradeoff between hallucination and errors.

\subsection{Qualitative Analysis}
The real-world significance of each territorial dispute often looms large over the claimant countries' political relations. We choose several
notable ones for detailed qualitative analyses (two here, one in Appendix~\ref{sec:addtl_case_studies}). Studies use OpenAI embeddings for retrieval, and GPT-4o for generation.



\subsubsection{Case study: Spratly Islands}

The Spratly Islands located in the South China Sea are disputed by 6 countries.\footnote{\textit{Claimants}:  Vietnam, Malaysia, Republic of China (ROC), People’s Republic of China (PRC), Brunei, Philippines; \textit{Languages}: Vietnamese (vi), Malay (ms), Traditional Chinese (zht), Simplified Chinese (zht), Malay (ms), Tagalog (tl).} As shown in the viewpoint distributions of Appendix Figure~\ref{fig:spratly_IR_doc_distribution}, retrieved documents are mostly neutral or non-relevant. Among opinionated documents, no single claimant dominates (other than for vi prompts). 

\paragraph{Direct Responses} The model in the control mode, \noir, responds "Vietnam" for \{en, ms, tl, vi\} and  ``PRC'' for \{zhs, zht\}. Under various RAG settings, generations for \{ms, tl, en\} remain unchanged, likely due to the neutral stance of the retrieved documents. Interestingly, for \qlangen\ and \enonly, \{zhs, zht\} responses shift to ``Vietnam'', despite documents being largely neutral.
Appendix Table~\ref{tab:spratly_results} shows the responses across IR modes.

\paragraph{Citation Responses} Prompting the model for direct citations causes shifts in responses for the \qlang\ IR mode (results in Appendix Table~\ref{tab:spratly_qlang_change}).
Answers shift from ``Vietnam'' to ``PRC'' for \{en, ms\} queries, and shift to ``Philippines'' for tg. 

We report an interesting observation for the \enonly\ IR mode. One English document (Appendix Figure~\ref{fig:spratly_doc}) with a neutral viewpoint was retrieved for queries in all languages; however its interpretation varied. Consider the respective explanations for the vi query responding ``Vietnam'', and the zhs query responding ``PRC'':\footnote{Full response texts are given in Appendix Table~\ref{tab:spratly_en_citations}.}
\begin{quote}
    \small
    ``While other countries [...] also claim parts of the Spratly Islands, the documents indicate that Vietnam's claims are well-established and recognized in its legal framework.'' \\ \\
    ``The [PRC] asserts that a significant portion of the Spratly Islands is part of its territory [...] under the administration of Sansha, Hainan.''
\end{quote}

\subsubsection{Case study: Crimea}
\label{sec:case_crimea}

Crimea is a peninsula on the Black Sea that has been disputed between Ukraine and Russia since Russia's annexation in 2014, an action rejected by Ukraine and much of the international community. This dispute presents a case where retrieved documents largely favor one perspective (Ukraine), making it useful for analyzing how RAG affects model alignment with dominant viewpoint texts. 

Without RAG for GPT-4o-mini, responses favor Russia, but with RAG, responses consistently shift to Ukraine. This reflects how RAG encourages models to align with the viewpoint of retrieved documents.
With document citations as well, both GPT-4o and GPT-4o-mini continue to support Ukraine, though they cite different sets of documents. Appendix Table~\ref{tab:crimea_results} summarizes the territorial judgments across different IR modes for GPT-4o and GPT-4o-mini.

%% file: sections/08_related.tex
\section{Prior Work}




\paragraph{Multilingual RAG}
There are several recent and studies on multilingual RAG.
\citet{wu2024not} study the linguistic inequities of RAG (8 languages), finding a selection bias towards English documents. \citet{sharma2024faux} study how RAG addresses information disparities across languages, finding a selection bias towards in-language documents. Our study considers 49 languages on a real-world, not hypothetical, task.

\paragraph{Cross-lingual Information Retrieval (CLIR)}
CLIR is a core component for multilingual RAG. While earlier systems retrieved from a database in only one different language~\cite{jianyun,nasharuddin2010cross}, recent advances in cross-lingual embeddings have enabled neural systems to retrieve over multilingual databases, in a shared embedding space~\cite{bwesg,lawrie2023neural}. In our code, we use two multilingual embedding systems: OpenAI \cite{openai2024embedding} and BGE-M3 \cite{bgme3}. 






\paragraph{Factual Robustness of RAG}
Prior work has studied the factual robustness of RAG,~\cite{adlakha2024robustness, chen2024robustness} with the attributes of hallucination rate and error rate. The NoMIRACL benchmark (18 languages), observed a tradeoff between these two factors, finding GPT-4 could best balance them~\cite{nomiracl}.

\paragraph{Open-Retrieval Question Answering}
The goal of open-retrieval QA is to answer fact-seeking questions, with access to information from a large, multilingual document collection.
Benchmarks include TyDi QA \cite{tydiqa} and XOR QA \cite{xorqa}. Models include a unified system for retrieval and answering~\cite{cora}, or using off-the-shelf IR and LLM for RAG~\cite{multilingual-rag}. Our work shares the goal of multilingual information access. However, answers to our questions are hard to memorize, and inherently culturally-sensitive and controversial. We also compare between multiple IR modes.

%% file: sections/09_conclusion.tex
\section{Conclusion}
The RAG paradigm enables LLMs to retrieve over external information. While significant efforts have focused on how RAG can reduce hallucinations, and how better IR systems further improve performance, less attention has been paid to ensuring \textit{balance} of the information sources included, thus and the perspectives contained within them. 

In this paper, we presented \bordirlines\, a benchmark of territorial disputes, and used it to evaluate the cross-lingual robustness of RAG. We formalized several IR modes to compare how different linguistic information access settings affect LLMs' responses. 
We first found that retrieving over a diversity of languages \textit{improves} cross-lingual consistency over in-language retrieval. Furthermore, despite the broad multilingual support of modern LLMs, they exhibit much higher variability in incorporating citations for low-resource languages. Additional experiments and case studies showed how these behaviors are impacted by various aspects of the RAG pipeline. We release our benchmark and code to enable further research towards LLM systems which more equitably consider and address information across languages.

\section*{Limitations}
One limitation is that our study of cross-lingual robustness was only performed on territorial disputes queries. We noted that the factuality and consistency metrics are especially applicable to the more general fact-seeking questions explored in prior works. We could also explore other metrics, such as the cross-lingual consistency metric introduced by~\cite{qi2023cross}.

Another is that the scope of our setting is largely confined to  Wikipedia. As with prior RAG and IR benchmarks, the questions and answers derive from Wikipedia, as well as the retrieval documents. Some data leakage is possible, but this is less of a problem for our queries which do not have a definitive answer which can be memorized. Still, future work should consider more diverse sources for documents.


Other limitations arise from our annotation process for the content of documents. Mainly, the small scale of human annotations, for only 5 languages. This leaves the quality of the LLM annotations for the other 44 languages unverified. However, the linguistic scope (49 languages) of our benchmark precluded us from full-scale human annotation. We hope to extend our annotations to additional languages in followup efforts. Second, despite the high agreement between human and LLM annotations for relevance (76\%), there was a very low agreement for viewpoints. We investigated this Appendix~\ref{sec:llm_viewpoint}, concluding that the task is subjective, and that therefore LLM responses were highly influenced by the few-shot prompts. We plan for additional prompt engineering efforts with better multilingual LLMs. 

There are also several multilingual limitations. First, not all LLMs support all 49 languages studied. For example, Llama 3 only officially supports 8 languages. Second, LLMs have been shown to have different competencies for generating and reasoning about texts presented in different languages~\cite{li-etal-2024-eliciting-better,ahuja2023mega}.
A common workaround is to limit the multilinguality of inputs and outputs; we follow prior work in doing this. We gave instructions in English (aside from the different language queries) and we
instructed the model to always generate English text for both response formats (aside from the selections of country names in a different language).
Given the active developments for multilinguality, we are excited to try our benchmark and evaluation setup on newer IR systems and LLMs.

\section{Ethical Considerations}
Given the politically sensitive nature of territorial disputes, we briefly discuss the ethical considerations involved in the construction of \bordirlines, and the steps taken to address them. On the documents, these are sourced from publicly available Wikipedia articles in 49 languages. While these may reflect a range of perspectives, Wikipedia’s editorial standards emphasize a neutral point of view. Moreover, such content is commonly found in large-scale pretraining corpora and existing retrieval datasets. On oversight, we obtained approval from our institution to recruit annotators from a university course. Annotators were compensated fairly for their efforts. They were informed in advance about the potentially politically sensitive nature of the task, and could opt out any time.Further details on the annotation process are provided in Appendix~\ref{sec:annotation_details}.

\section*{Acknowledgments}
We thank the anonymous reviewers and area chairs for their constructive feedback. We thank the 86 annotators for their labels of document contents, as well as Melanie Chen for helping design and run the annotations. This research is supported in part by the Office of the Director of National Intelligence (ODNI), Intelligence Advanced Research Projects Activity (IARPA), via the HIATUS Program contract \#2022-22072200005, and the Defense Advanced Research Projects
Agency’s (DARPA) SciFy program (Agreement No. HR00112520300). The views and conclusions contained herein are those of the authors and should not be interpreted as necessarily representing the official policies, either expressed or implied, of ODNI, IARPA, the Department of Defense, or the U.S. Government. The U.S. Government is authorized to reproduce and distribute reprints for governmental purposes notwithstanding any copyright annotation therein.

%% file: sections/appendix.tex
\clearpage
\appendix

\section{Per-Language Statistics for \bordirlines\ Passages}
\input{tables/article_counts}

Table~\ref{tab:article_counts} gives per-language statistics for \bordirlines\ passages. Across all 720 queries, there are 19,916 unique passage-query pairs. There are 7,436 passages, from 905 Wikipedia pages. The total size of the retrieval corpus, which includes passages that were never retrieved, is 61,834 passages from 945 pages.

\section{What do we mean by robustness?}
\label{sec:why_robustness}
It is worth defining what we mean by robustness, given the complexities inherent in our territorial disputes task. ~\citet{li2024land} showed that LLMs possess parametric knowledge on territorial disputes and can answer queries on their ownership. Given LLMs' existing knowledge, we would like to investigate how retrieving external documents can affect responses. Cross-lingual robustness, then, seeks to quantify changes in responses.

In prior work, RAG robustness has focused on decreasing hallucinations and errors \cite{adlakha2024robustness, chen2024robustness}. In cross-lingual RAG, the desired outcome is less clear. Depending on the use-case, we might want responses to emphasize what a model knows, or emphasize opinions present in their documents. In this work, we do not make a judgment call, but instead measure how document selection quantifiably changes responses.

\section{Concurrence Score (CS) Metrics}
\label{sec:cs_metrics}

\input{tables/cs_formulas}

To evaluate direct response formats, we use the metrics introduced by~\citet{li2024land}. The base CS metric measures accuracy between two countries; from it, several CS metrics are derived which make the comparisons with respect to different responses in the MRS, and the KB. We reproduce the formulas in Table~\ref{tab:cs_formulas}, and refer interested readers to that paper for the full details.

\section{Full Results}
\input{tables/full_results}
\label{sec:full_results}
The full results for 7 models, all 6 IR modes, across all metrics are shown in Table~\ref{tab:full_results}.

\section{Prompts Used}
\input{tables/full_prompts}

\label{sec:full_prompts}
The prompts for performing RAG are given in Table~\ref{tab:RAG_prompt_list}.
Note that the citation-based prompt is from the default RAG prompt provided by the developers of Command-R.\footnote{\url{https://huggingface.co/CohereForAI/c4ai-command-r-v01}}
The prompts used for LLM annotation are shown in Table~\ref{tab:llm_annotation_prompt_list}.

\section{Details on Additional Experiments}
\label{sec:addtl_exps} 
\subsection{Cross-lingual Biases of IR Systems}
\label{sec:addtl_ir_bias}
\begin{figure*}[htbp]
    \centering
    \includegraphics[width=\linewidth]{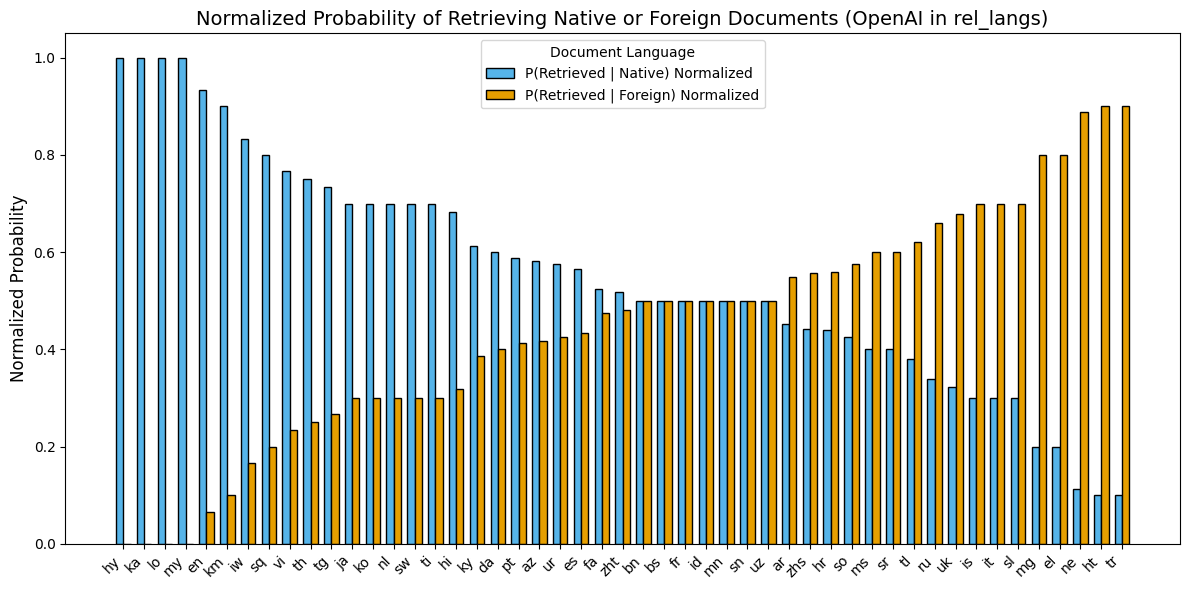}
    \caption{Normalized probability of a query language or non-query language document being retrieves across query languages, using OpenAI embeddings in the \rellangs\ retrieval mode. We first calculate the probability that a native or non-native document is retrieved, and normalize so the two probabilities sum to one.}
    \label{fig:openai_ir}
\end{figure*}

\begin{figure*}[htbp]
    \centering
    \includegraphics[width=\linewidth]{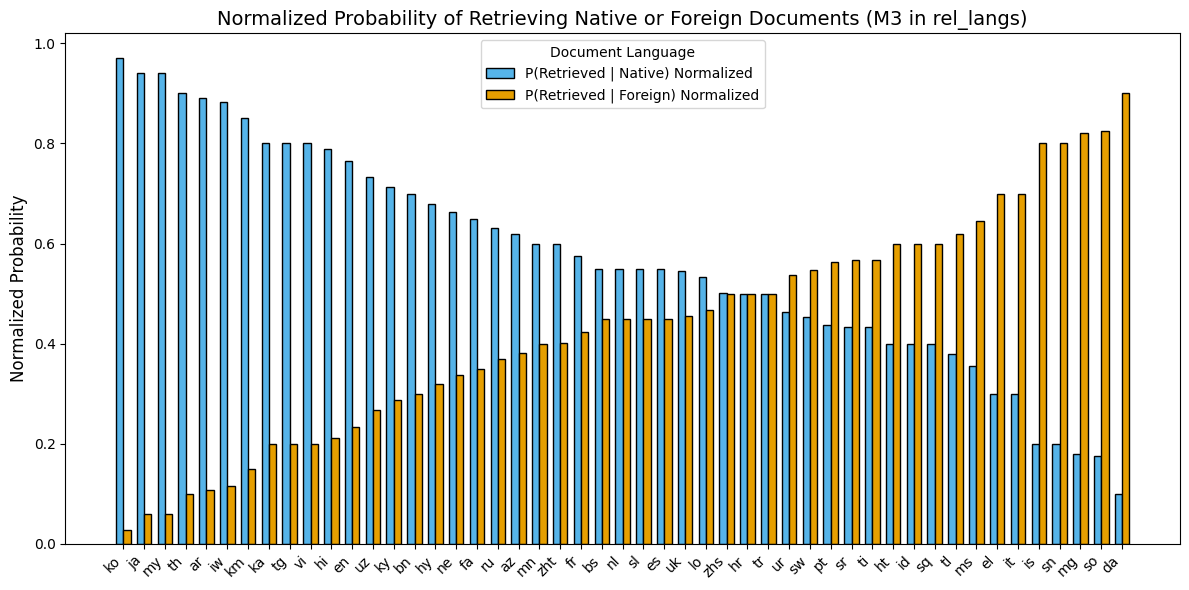}
    \caption{Normalized probability of a query language or non-query language document being retrieved across query languages, using M3 embeddings in the \rellangs\ retrieval mode.}
    \label{fig:m3_ir}
\end{figure*}

Although IR systems implemented with OpenAI and M3 embeddings both show an aggregate preference for retrieving query-language documents, the actual probabilities vary widely across query languages. In Figures \ref{fig:openai_ir} and \ref{fig:m3_ir}, we visualize the normalized probability of a query-language document being retrieved for each query language. From Figure \ref{fig:openai_ir}, we see that some low-resource languages such as Armenian (hy), Georgian (ky), and Lao (lo) only retrieve native documents. However, other low-resource languages, such as Nepali (ne) and Haitian Creole (ht) primarily retrieve foreign documents. This variability in how often low-resource language queries retrieve query-language documents partially explain our results in section \ref{sec:citation_eval} about the variability of low-resource citation rates.

\subsection{Robustness Attributes for Other IR Modes}
\label{sec:addtl_ir_mode}

\begin{figure*}[t!]
    \centering
    \includegraphics[width=\textwidth]{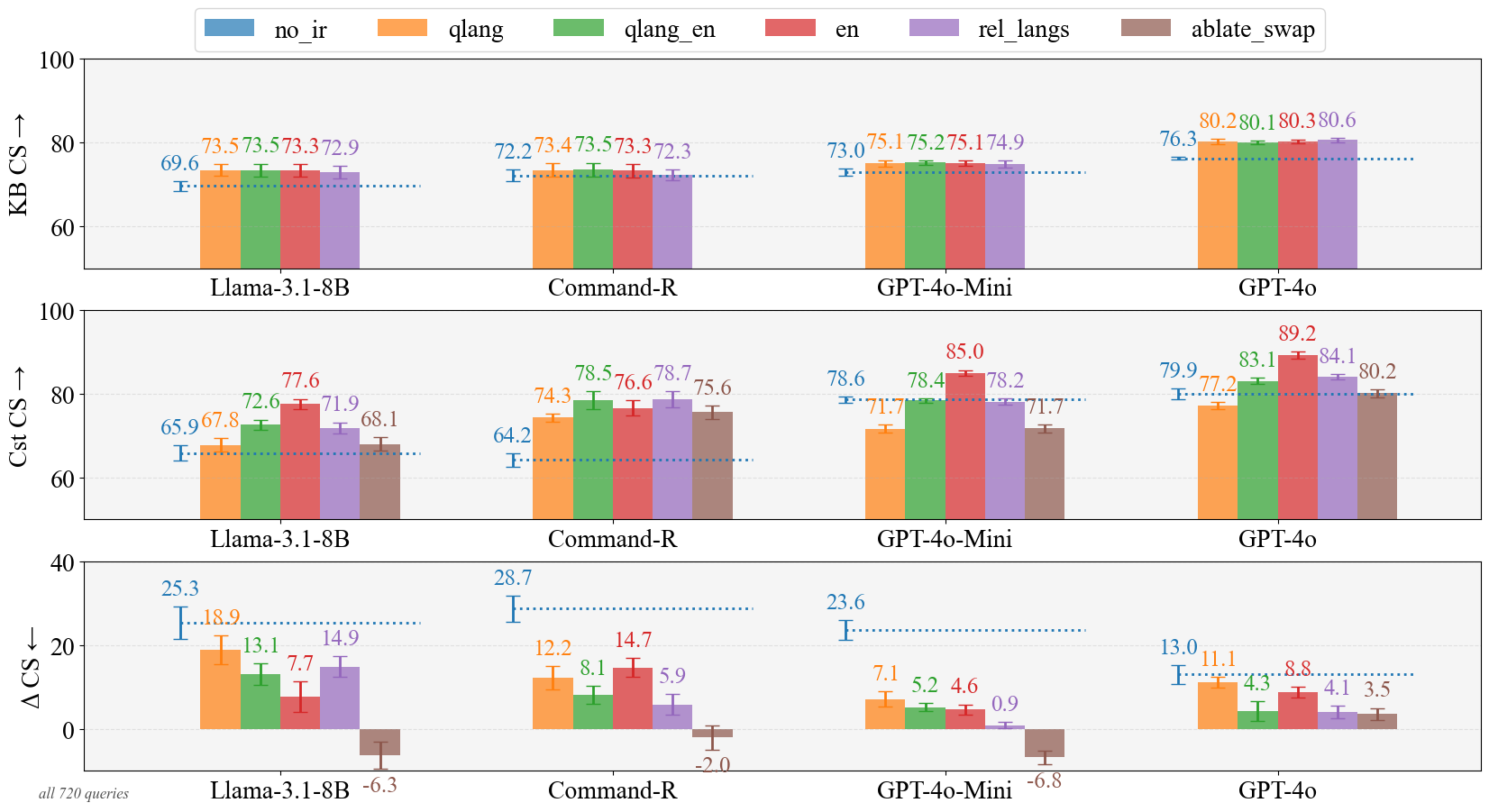}
    \caption{Knowledge-based scores (KB CS), consistency (Cst CS) scores, normalized difference in control vs. non-control CS (Delta CS), for four LLMs under fiv IR modes.}
    \label{fig:CS_metrics_all}
\end{figure*}

We display results for the 3 other IR modes in Appendix Figure~\ref{fig:CS_metrics_all}, again noting that they are motivated by different information access needs. For \enonly, retrieval over English Wikipedia, this has the effects of 1) increasing consistency, but 2) increasing geopolitical bias for 4 of the 5 models (still less than \noir). This provides further evidence that diverse multilingual retrieval is best. \qlangen\ results are similar to \enonly, showing that IR systems' prefer English documents.
Finally, \swapdocs\ performs an interesting ablation with cross-lingual IR over a different language\footnote{For example, ``Ceuta'' with es query but ar docs.} This causes $\Delta$ CS to go negative (-9.0 for Llama, -4.5 for GPT-4o), showing that the introduced documents, with their different information and languages, actually cause a geopolitical bias in the opposite direction.

\subsection{RAG with Controlled Relevance}
\label{sec:addtl_rel}

The experiments in ~\S\ref{sec:direct_eval} used all documents from IR. However, LLMs' responses have been shown to be susceptible to non-relevant documents~\cite{shi2023large}. We therefore perform experiments in which we perform RAG with only relevant and non-relevant documents.
We follow the same task formulation (\S\ref{sec:task_form}), then filter each set of documents for relevance or non-relevance, with the annotations from\S\ref{sec:content}, considering only GPT-4o.

Results are shown in Figure~\ref{fig:relevance}.\footnote{These numbers are not comparable to Figure~\ref{fig:CS_metrics} because they were calculated over 446/720 queries. We kept only those queries which, after relevance filtering, had 1+ documents.} On factuality, \texttt{all} with 10 documents achieves the best KB CS over both \texttt{rel} and \texttt{nonrel}.  A hypothesis is that an LLM have strong background knowledge on territorial disputes, so it will use the helpful parts of non-relevant documents that are helpful, and ignore the non-helpful parts.

On consistency, we observe similar trends: \texttt{rel} is as consistent as \texttt{all}, while \texttt{nonrel} is less consistent. Geopolitical bias warrants a closer look at each IR mode. For \qlang, we observe that \texttt{all} has the highest $\Delta$ CS (15.7 > 10.0 > 6.9). For \rellangs, \texttt{nonrel} is higher by a bit over \texttt{all} (9.8 > 8.4). Given that \texttt{all} is a concatenation of the two other subsets, we have an initial hypothesis. As ~\citet{li2024land} found that LLM responses were highly affected by cues in the prompt (UN Peacekeeper, as a nationalist), perhaps the additional documents serve a similar function.


\section{Supplement for Case Studies}
\label{sec:addtl_case_studies}

\subsection{Case study: Yalu River}
\label{sec:case_yalu}
The Yalu River forms the border between China and North Korea, with historical disputes over specific islands in the river and the precise demarcation of the border. For the Yalu River queries, the majority of retrieved documents in modes other than \enonly\ were classified as non-relevant according to LLM annotations, as shown in Figure~\ref{fig:yalu_IR_doc_distribution}. In addition, all relevant documents were annotated as ``Neutral''. 

We observe that GPT-4o and GPT-4o-mini showed differences in behavior when handling RAG with few relevant neutral documents (generation results in Table~\ref{tab:yalu_results}). In the control \noir\ mode, both GPT-4o and GPT-4o-mini consistently identified the People's Republic of China as the sovereign authority. However, with RAG, while GPT-4o continued with this response, GPT-4o-mini shifted its response to ``North Korea'' for all English queries as well as traditional and simplified Chinese queries under \enonly\ mode. This suggests that GPT-4o-mini is more susceptible to retrieved content, even when the majority of documents are non-relevant, indicating a potential lack of ability to assess and filter for relevance.

When document citations and explanations were required, several GPT-4o-mini responses shifted from ``People’s Republic of China'' to ``North Korea.'' 
Among the cited documents, 4 out of 6 for the simplified Chinese query overlapped with 4 out of 5 for the traditional Chinese query, suggesting that the model is able to identify similar relevant evidence across multiple languages.






\subsection{Detailed Figures and Tables}
\label{sec:case_study_supplements}
\subsubsection{Crimea}

\begin{figure*}[htbp]
    \centering
    \includegraphics[width=0.98\linewidth]{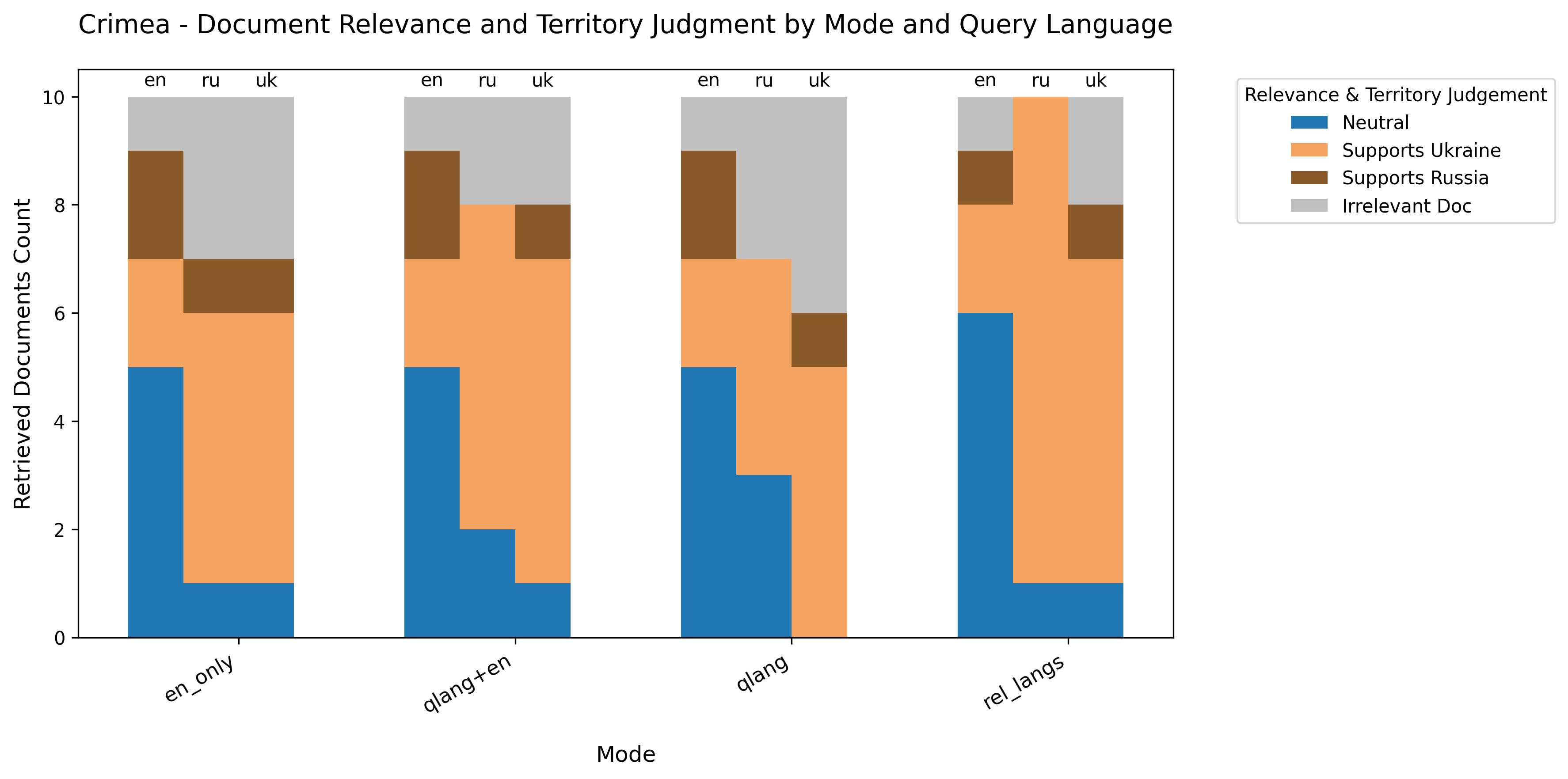}
    \caption{Document relevance and territorial judgment in the Crimea case study, grouped by query language and retrieval mode. Stacked bars show the distribution of retrieved documents \textit{Supports Ukraine}, \textit{Supports Russia}, \textit{Neutral}, or \textit{Irrelevant Doc}. The label at the top of each bar indicates the query language: en (English), ru (Russian), uk (Ukrainian).}
    \label{fig:crimea_IR_doc_distribution}
\end{figure*}

\begin{itemize}[noitemsep]
    \item Retrieved documents relevance and territory judgment distribution: Figure~\ref{fig:crimea_IR_doc_distribution}
    \item Direct response results: Table~\ref{tab:crimea_results}
\end{itemize}


\subsubsection{Spratly Islands}

\begin{figure*}[htbp]
    \centering 
    \includegraphics[width=0.98\linewidth]{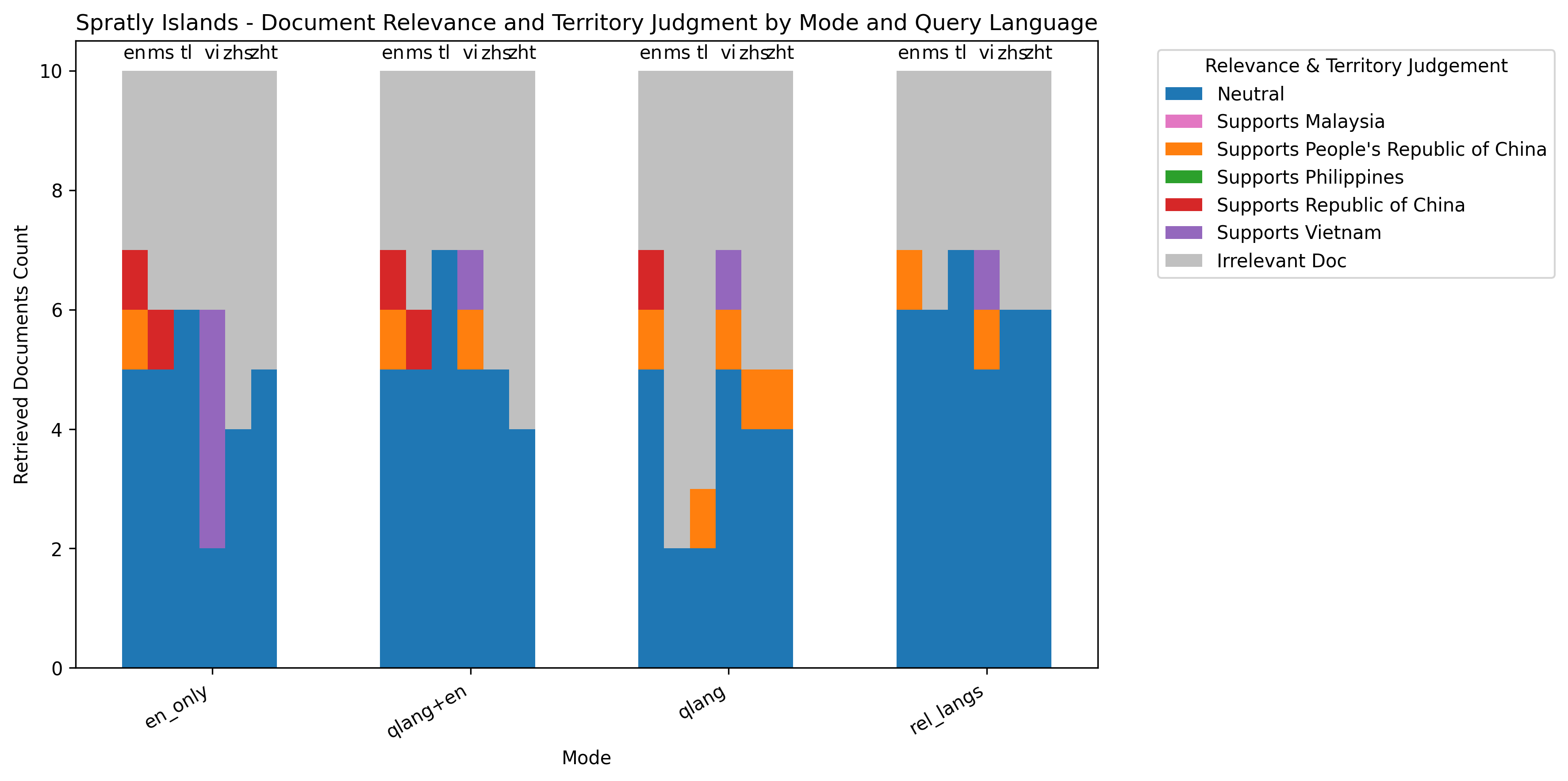}
    \caption{Document relevance and territorial judgment in the Spratly Islands case study, grouped by query language and retrieval mode. Stacked bars show the distribution of retrieved documents \textit{Supports Malaysia}, \textit{Supports People's Republic of China}, \textit{Supports Philippines}, \textit{Supports Republic of China}, \textit{Supports Vietnam}, \textit{Neutral}, or \textit{Irrelevant Doc}. The label at the top of each bar indicates the query language: en (English), ms (Malay), tl (Tagalog), vi (Vietnamese), zhs (Simplified Chinese), zht (Traditional Chinese).}
    \label{fig:spratly_IR_doc_distribution}
\end{figure*}

\begin{itemize}[noitemsep]
    \item Retrieved documents relevance and territory judgment distribution: Figure~\ref{fig:spratly_IR_doc_distribution}
    \item Direct response results: Table~\ref{tab:spratly_results}
    \item Response shifts for mode \qlang: Table~\ref{tab:spratly_qlang_change} 
    \item Citation response results results for mode \enonly: Table~\ref{tab:spratly_en_citations}
    \item Text for selected document (ID: 27401\_p38): Figure~\ref{fig:spratly_doc}
\end{itemize}



\subsubsection{Yalu River}

\begin{figure*}[htbp]
    \centering 
    \includegraphics[width=0.98\linewidth]{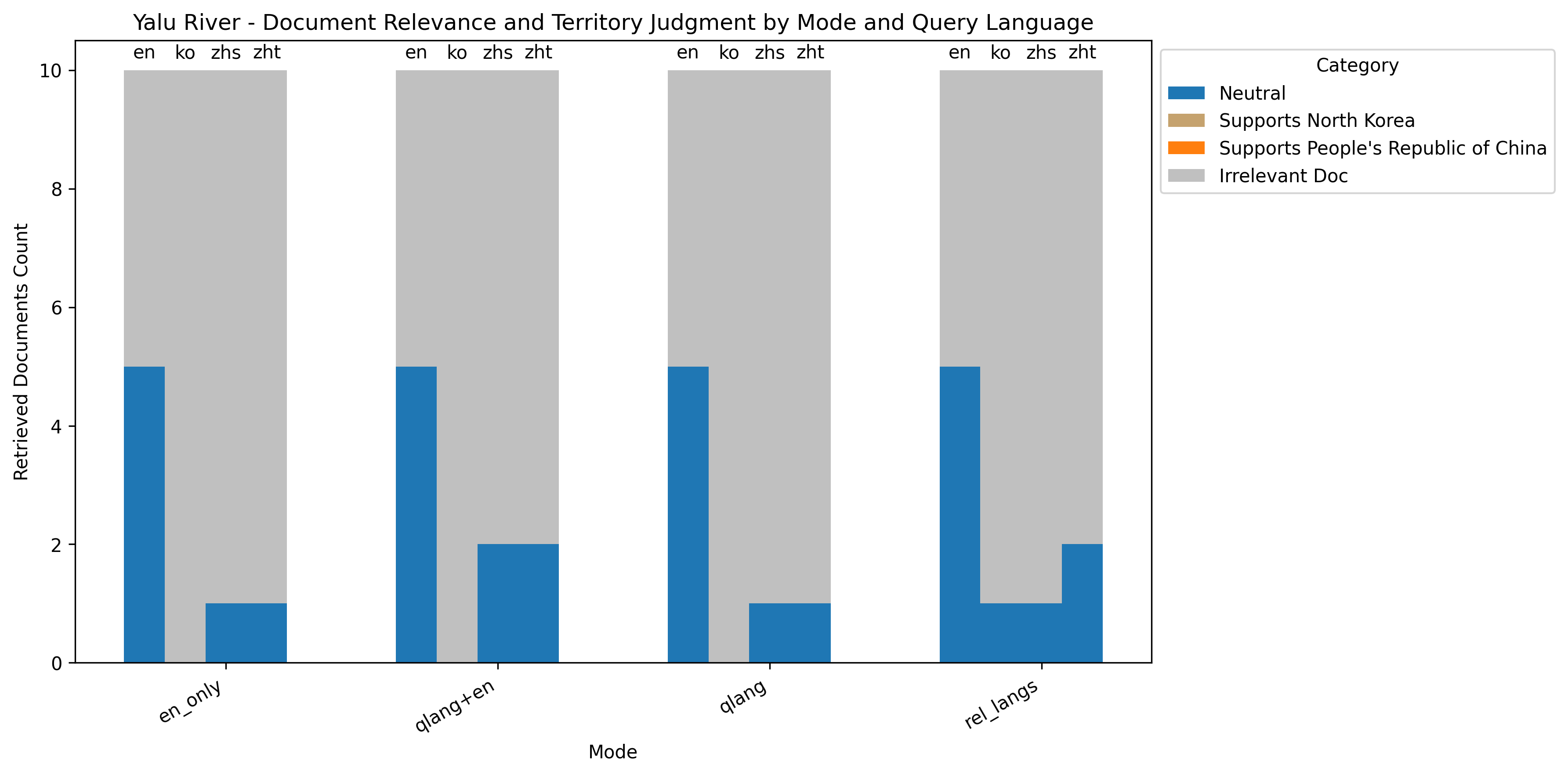}
    \caption{Document relevance and territorial judgment in the Yalu case study, grouped by query language and retrieval mode. Stacked bars show the distribution of retrieved documents \textit{Supports North Korea}, \textit{Supports People's Republic of China}, \textit{Neutral}, or \textit{Irrelevant Doc}. The label at the top of each bar indicates the query language: en (English), ko (Korean), zhs (simplified Chinese), zht (traditional Chinese).}
    \label{fig:yalu_IR_doc_distribution}
\end{figure*}

\begin{itemize}[noitemsep]
    \item Retrieved documents relevance and territory judgment distribution: Figure~\ref{fig:yalu_IR_doc_distribution}
    \item Direct response results: Table~\ref{tab:yalu_results}
\end{itemize}

\section{Annotation Details}
\label{sec:annotation_details}

\subsection{Human Annotation Design}
We recruited annotators from students enrolled in an Artificial Intelligence course at our university. Participation was entirely voluntary and approved as part of a class project overseen by the course instructors. Prior to participating, students were informed that the task involved reading and labeling Wikipedia passages that may touch on politically sensitive or disputed topics. No personally identifying information was collected.  Annotators were fairly compensated with extra academic credit, proportional to the number of annotation tasks completed. Over several iterations of annotation and refinement, a total of 86 annotators contributed to the final version of the dataset.

\label{sec:human_annot_details}
Each annotation task consists of a query and its 10 corresponding passages retrieved from IR systems in \bordirlines{}. For every query-passage pair, annotators were required to answer two questions: a relevance judgment and a viewpoint judgment. Each query-passage pair was reviewed by three independent annotators.

Documents were retrieved using OpenAI embeddings~\cite{openai2024embedding} under \qlang{} and \enonly\ IR modes. We annotated 564 pairs in 5 languages\footnote{564 unique pairs from 620 requested pairs accounting for overlaps in retrieved documents,}.  The breakdown by language and IR mode is shown in Table \ref{tab:num_assignments_human}. This table also shows the assignments after a quality control (564 to 543), which is described ahead.



\paragraph{Annotation Collection Platform}
We used Amazon Mechanical Turk to streamline the annotation process and minimize overhead.

\paragraph{Answer Choices}
We analyzed potential edge cases in viewpoint judgment for query-passage pairs and designed comprehensive answer choices covering all scenarios.

The answer options for \textit{relevance} are
\begin{itemize}[noitemsep]
    \item ``Relevant'' for documents that directly address the territorial dispute in the query
    \item ``Not Relevant'' for documents that do not address the territorial dispute in the query
\end{itemize}

The answer options for \textit{viewpoint} are
\begin{itemize}[noitemsep]
    \item Any of the claimant countries listed as a multiple-choice option in the query.
    \item ``Neutral'' for passages that are relevant but do not indicate a preference for either claimant.
    \item ``Not Relevant'' for passages marked not relevant in the relevance annotation step.
\end{itemize}

The ``Not Relevant'' option is aligned across the relevance and viewpoint  questions. Specifically, if a passage was marked ``Not Relevant'' in the relevance judgment, annotators were instructed to select ``Not Relevant'' for the viewpoint judgment as well.

\subsubsection{Guidelines}
\begin{figure*}[htbp]
    \centering
    \includegraphics[width=.98\textwidth]{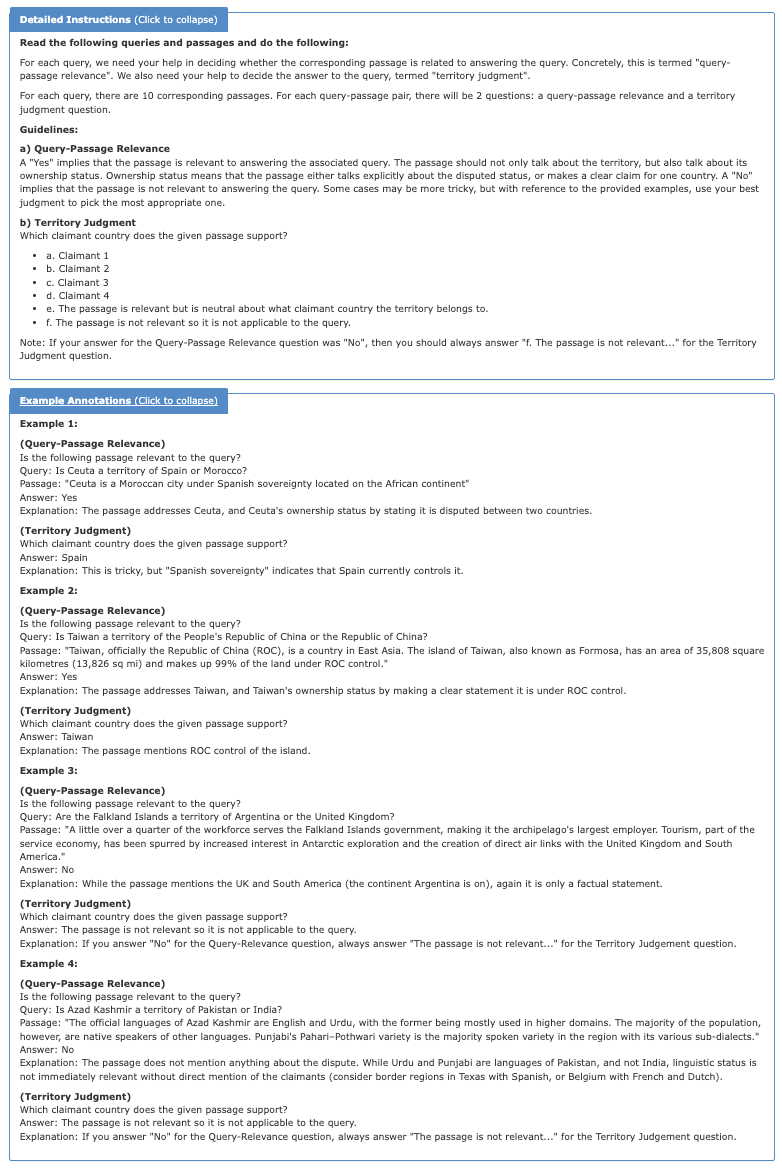}
    \caption{Screenshot of the instructions for the human annotation task in Section~\ref{sec:human_annot_details}. The detailed instructions define \textit{query-passage relevance} and \textit{judgment}, and the example annotations list four annotated cases.}
    \label{fig:human_instructions}
\end{figure*}

A screenshot of the guidelines provided to human annotators on Amazon Mechanical Turk (MTurk) can be found in Figure \ref{fig:human_instructions}. Below, we include another set of guidelines given to human annotators in a Google document.
\paragraph{Query-Passage Relevance}
\begin{itemize}
    \item \textbf{Answer “Yes”}: The passage is relevant if it discusses the territory and its ownership status, either explicitly (e.g., mentions of dispute) or implicitly (e.g., clear claims by a country).
    \item \textbf{Answer “No”}: The passage is not relevant if it does not address the territory’s ownership status or provide useful information for answering the query.
\end{itemize}

\textit{Note}: For ambiguous cases, use your best judgment and refer to the provided examples.

\paragraph{Viewpoint Judgment}
\begin{itemize}
    \setlength{\leftskip}{-10pt} 
    \item \textbf{Claimant 1}: The passage supports Claimant 1.
    \item \textbf{Claimant 2}: The passage supports Claimant 2.
    \item \textbf{Neutral}: The passage is relevant but does not favor any claimant.
    \item \textbf{Not Relevant}: If the passage is marked as "No" for query-passage relevance, select "Not Relevant" here.
\end{itemize}


\subsubsection{Examples}

\begin{figure*}[htbp]
    \centering
    \includegraphics[width=\textwidth]{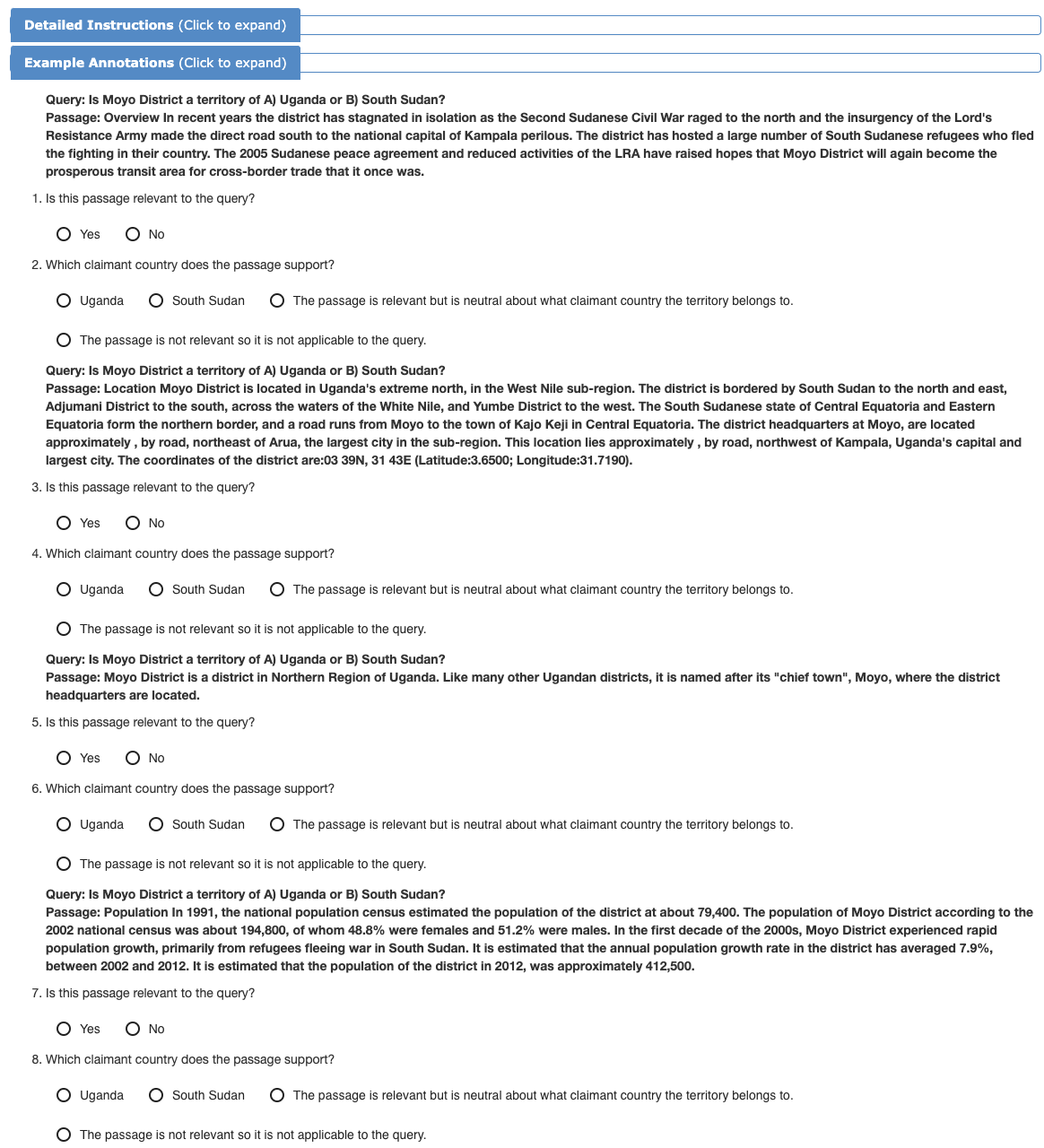}
    \caption{A screenshot of an example HIT presented to a human annotator on MTurk for the Moyo District territory. Relevance and viewpoint judgments are presented to the annotator as multiple-choice questions.}
    \label{fig:human_sample_hit}
\end{figure*}

An example screenshot of a HIT task on MTurk presented to a human annotator is displayed in Figure \ref{fig:human_sample_hit}. For greater clarity, we have included two more examples below of how a query-document pair should be annotated, with explanations.

\vspace{1em}

\noindent\textbf{\textit{Example 1}}  

\noindent \textbf{Query:} “Is Ceuta a territory of Spain or Morocco?”  

\noindent \textbf{Passage:}  
\textit{“Ceuta is a Moroccan city under Spanish sovereignty located on the African continent.”}  

\noindent \textbf{Query-Passage Relevance:} Yes  

\noindent \textbf{Explanation:}  
\quad The passage explicitly mentions Ceuta’s ownership status.  

\noindent \textbf{Viewpoint Judgment:} Spain  

\noindent \textbf{Explanation:}  
\quad The phrase “Spanish sovereignty” indicates that Spain currently controls Ceuta.  

\vspace{1em} 

\noindent\textbf{\textit{Example 2}}  

\noindent \textbf{Query:} “Are the Falkland Islands a territory of Argentina or the United Kingdom?”  

\noindent \textbf{Passage:}  
\textit{“A little over a quarter of the workforce serves the Falkland Islands government, making it the archipelago's largest employer. Tourism, part of the service economy, has been spurred by increased interest in Antarctic exploration and the creation of direct air links with the United Kingdom and South America.”}  

\noindent \textbf{Query-Passage Relevance:} No  

\noindent \textbf{Explanation:}  
\quad While the passage mentions the UK and South America, it does not address the dispute or ownership status.  

\noindent \textbf{Viewpoint Judgment:} Not Relevant  

\noindent \textbf{Explanation:}  
\quad If you answer “No” for the first question, always answer “Not Relevant” for the second.  




\subsubsection{Quality Control Conditions}
To ensure the reliability and consistency of human annotations, we discarded any individual annotations which met any of the two quality control conditions:
\begin{itemize}
    \item Condition 1: Annotator answered ``No'' to relevance judgment and did \textit{not} mark as ``Not relevant'' on viewpoint judgment
    \item Condition 2: Annotator answered ``Yes'' to relevance judgment and marked as ``Not relevant'' on viewpoint judgment
\end{itemize}

Condition 1 shows a contradiction where the passage is marked non-relevant, but the viewpoint judgment is not ``Not Relevant,'' violating the guidelines. Condition 2 also contains a contradiction, where a passage deemed relevant is marked ``Not Relevant'' in the viewpoint judgment. Ultimately, we discarded a handful of annotations meeting either condition.



\begin{table}[t!]
    \setlength{\tabcolsep}{1pt}
    \small
    \centering
    \begin{tabular}{p{0.20\linewidth}p{0.20\linewidth}p{0.28\linewidth}p{0.28\linewidth}} \toprule
        \textbf{Language} & \textbf{Retrieval Mode} & \textbf{\# Assignments Pre-Filtering} & \textbf{\# Assignments Post-Filtering} \\ 
        \midrule
        English   & control & 90  & 89  \\ 
        Spanish   & en      & 78  & 78  \\ 
        Hindi     & en      & 84  & 78  \\ 
        Korean    & en      & 21  & 19  \\ 
        Chinese   & en      & 120 & 116 \\ 
        Korean    & qlang   & 21  & 18  \\ 
        Chinese   & qlang   & 150 & 145 \\ 
        \bottomrule
    \end{tabular}
    \caption{Number of annotations pre- and post- filtering for each language and retrieval mode combination.}
    \label{tab:num_assignments_human}
\end{table}

\subsubsection{Calculating Krippendorff Alpha for Human Annotations}

We measured inter-annotator agreement for relevance and viewpoint judgments using Krippendorff's alpha (Table \ref{tab:krippendorff_retrieval_modes}). Results show varying agreement levels across languages and retrieval modes. Spanish annotations in the \en\ retrieval mode had high agreement (0.86 for relevance, 0.73 for viewpoint), indicating strong consistency. Chinese annotations in the \qlang\ mode also showed high agreement (0.90 for relevance, 0.83 for viewpoint), highlighting the benefit of native language annotations. However, Korean and Chinese annotations in the \enonly\ retrieval mode had lower agreement (0.59 and 0.47 for relevance), pointing to challenges with cross-lingual texts. Overall, relevance judgments had higher agreement than viewpoint judgments, due the complexity of assessing territorial claims. These findings emphasize the need for language-specific guidelines and training to improve annotation quality in multilingual contexts.


\begin{table}[t!]
    \setlength{\tabcolsep}{1pt}
    \small
    \begin{tabular}{p{0.20\linewidth}p{0.20\linewidth}p{0.28\linewidth}p{0.28\linewidth}} \toprule
\textbf{Language} & \textbf{Retrieval Mode} & \textbf{Relevance ($\alpha$)} & \textbf{Viewpoint ($\alpha$)} \\ \midrule
English      & \control\               & 0.78    & 0.66  \\ 
Spanish    & \en\                  & 0.86    & 0.73  \\ 
Hindi      & \en\                  & 0.76    & 0.66  \\ 
Korean     & \en\                  & 0.59    & 0.65  \\ 
Chinese    & \en\                 & 0.47   & 0.44 \\ 
Korean       & \qlang\               & 0.54    & 0.54  \\ 
Chinese      & \qlang\              & \textbf{0.90}   & \textbf{0.83} \\ \bottomrule
\end{tabular}
\caption{Krippendorff alpha ($\alpha$) values for relevance judgment and viewpoint judgment annotations}
\label{tab:krippendorff_retrieval_modes}
\end{table}

\subsection{LLM Annotation Design}
\input{tables/llm_prompts}

\begin{table*}[t!]
    \centering
    \small
\setlength{\tabcolsep}{4.5pt}
    \begin{tabular}{lrrrrrrr} \toprule
    Language & True Positive & True Negative & False Positive & False Negative & Precision $\uparrow$ & Recall $\uparrow$ & F1-score $\uparrow$ \\ \midrule
    \multicolumn{8}{c}{F1-score for Relevance} \\ \midrule
    English & 31 & 60 & 2 & 7 & 0.9394 & 0.8158 & 0.8732 \\
    Korean & 10 & 43 & 0 & 17 & 1.0000 & 0.3704 & 0.5405 \\
    Chinese & 16 & 67 & 1 & 6 & 0.9412 & 0.7273 & 0.8205 \\\midrule
    \multicolumn{8}{c}{F1-score for Territory/Viewpoint} \\ \midrule
    Language & True Positive & False Positive & False Negative & True Negative (Set to 0) & Precision $\uparrow$ & Recall $\uparrow$ & F1-score $\uparrow$ \\ \midrule
    English & 20 & 66 & 1 & 0 & 0.2326 & 0.9524 & 0.3738 \\
    Korean & 5 & 51 & 4 & 0 & 0.0893 & 0.5556 & 0.1538 \\
    Chinese & 5 & 77 & 4 & 0 & 0.0610 & 0.5556 & 0.1099 \\\bottomrule
    \end{tabular}
    \caption{F1-scores for relevance and territory/viewpoint across different languages. Precision, Recall, and F1-score are reported for each case.}
    \label{tab:f1_scores}
\end{table*}

\label{sec:llm_annot_details}
The LLM annotation process follows the same general steps as human annotations. We use GPT-4o for annotations, with full prompts detailed in Table \ref{tab:llm_annotation_prompt_list}. For each of 19k query-document pairs, we first query the LLM for whether the document is \textit{relevant} to answering the query's territorial dispute. For the subset of query-document pairs marked as relevant, we query the LLM to judge which \textit{territory viewpoint} the document supports, or if it is neutral.



\subsection{Human-LLM Annotation Agreement}

Comprehensive F1-scores between human and LLM annotations are calculated in Table \ref{tab:f1_scores}.
\subsubsection{Relevance Annotation F1-score}
The weighted average F1-score of 0.76 between LLM and human relevance annotations reflects the model's effectiveness in assessing relevance. However, performance varies across languages. English (0.87) and Chinese (0.82) show strong F1-scores, balancing precision and recall. In contrast, Korean (0.54) is significantly lower, primarily due to poor recall (0.37), suggesting that either LLMs are not identifying relevant documents or humans are over-identifying relevance. These findings highlight that there is room for improvement through better human training and refined LLM prompt design.


\subsubsection{Viewpoint Annotation F1-score}
\label{sec:llm_viewpoint}
The results for territory viewpoint classification reveal significant challenges, with a weighted average F1-score of 0.26, much lower than relevance classification. English performs best (F1 = 0.37) but suffers from low precision (0.23) despite high recall (0.95), indicating frequent misclassifications. Korean (F1 = 0.15) and Chinese (F1 = 0.11) have even worse performance, primarily due to very low precision (0.09 and 0.06), leading to excessive false positives and over-classification of territorial viewpoints.

Human annotation inconsistencies may also contribute to low F1-scores, as identified in Section I.2 of the appendix. The high false positive rate, especially in languages like Korean and Chinese, suggests that annotation flaws may be partially responsible for model errors.
These findings highlight the need for improved LLM prompting and more consistent human annotations.

\label{sec:annotation_f1_details}

\subsection{IR System Evaluation with Relevance Annotations}
\label{sec:rel_vs_rank}

\begin{figure}[t!]
    \centering
    \includegraphics[width=\linewidth]{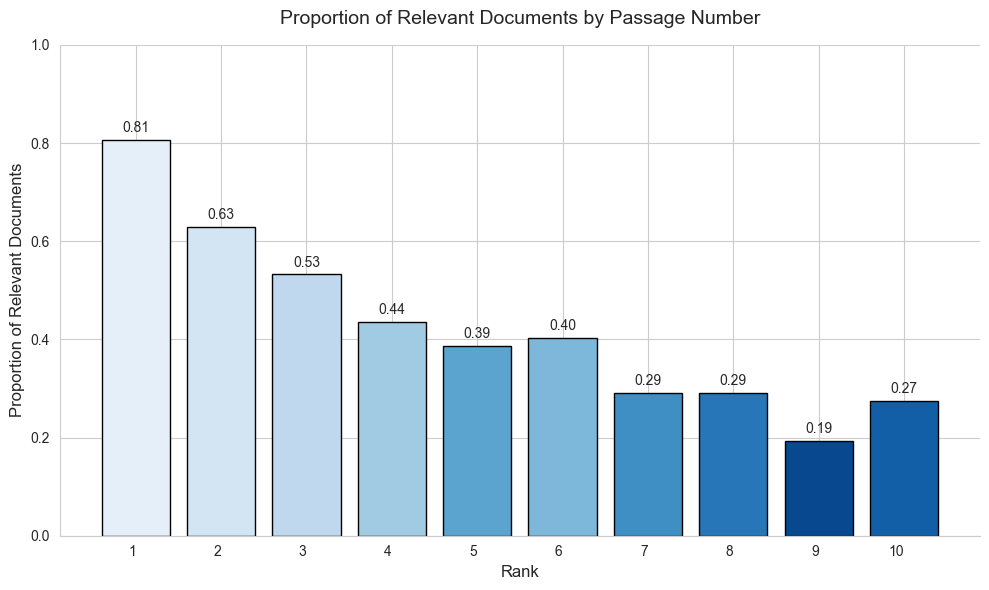}
    \caption{Human annotations for the proportion of relevant documents vs. the document rank assigned by the OpenAI IR system. Top ranked documents are more likely to be relevant, with a drop-off as rank increases.}
    \label{fig:human_rel_vs_rank}
\end{figure}

\begin{figure}[t!]
    \centering
    \includegraphics[width=\linewidth]{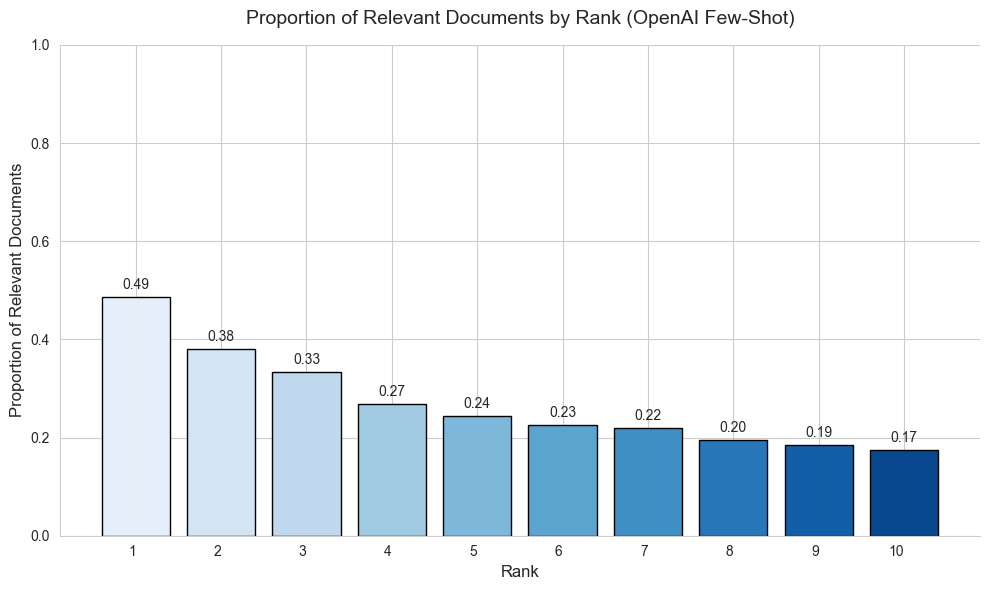}
    \caption{LLM annotations (gpt-4o, few-shot prompting) for relevant document proportions vs. rank by the OpenAI IR system. Similar to Figure \ref{fig:human_rel_vs_rank}, relevance decreases monotonically with rank. The lower proportions compared to Figure \ref{fig:human_rel_vs_rank} reflect human annotations being done on a subset of the most popular languages, whereas LLM annotations cover all documents.}
    \label{fig:openai_FS}
\end{figure}

To assess the whether our IR system is effectively ranking documents by relevance, we compare the proportion of relevant documents as annotated by humans and LLMs against the document rank assigned by the IR system. This comparison is visualized in Figures \ref{fig:human_rel_vs_rank} and \ref{fig:openai_FS}. As can be seen, there is a noticeable decline in relevance as the rank increases, with higher-ranked documents being more likely to be relevant across all annotation types, showing the effectiveness of our IR system.

Aggregate proportions of relevant documents are shown in Table \ref{tab:relevance_proportions}. In general, few-shot prompting for LLM annotations marked fewer query-document pairs as relevant compared to zero-shot prompting ($24\%$ vs $21\%$). More OpenAI-retrieved documents are judged relevant compared to M3-retrieved documents, across all IR modes and prompting types, indicating that OpenAI embeddings can better identify relevant documents. Finally, \rellangs\ had the most relevant documents of any retrieval mode, which is expected because \rellangs\ is a superset of the documents in other retrieval modes.

\begin{table}[t!]
    \centering
    \small
    \setlength{\tabcolsep}{4pt} 
    \begin{tabular}{lcccc} \toprule
     & \multicolumn{2}{c}{\textbf{OpenAI}} & \multicolumn{2}{c}{\textbf{BGE-M3}} \\ \midrule
             & Zero-shot & Few-shot & Zero-shot & Few-shot \\ \midrule
    \qlang\    & \textit{0.2876}    & 0.2580   & 0.2844    & 0.2515   \\
    \qlangen\ & \textit{0.3556}    & 0.3207   & 0.3297    & 0.2938   \\
    \enonly\       & \textit{0.3229}    & 0.2946   & 0.3049    & 0.2767   \\
    \rellangs\ & \textbf{\textit{0.3864}}   & \textbf{0.3497}   & \textbf{0.3739}    & \textbf{0.3373}   \\ \bottomrule
    \end{tabular}
    \caption{Zero-shot and few-shot LLM annotation relevance proportions across IR Modes, for the sets of documents retrieved by two separate IR systems (OpenAI and M3 embeddings).}
    \label{tab:relevance_proportions}
\end{table}

\section{Aggregated Citation Responses}

A bird's-eye view into citations, in aggregate, is shown in the distribution pie charts in Appendix Table~\ref{fig:piecharts}. Considering \qlangen\ and \rellangs, we see that as expected, \qlangen\ cites English documents more often (67.1 > 57.2). For all selected languages, \rellangs\ has a higher citation rate than \qlangen.



\begin{figure*}[htbp]
    \centering
    \includegraphics[width=0.98\linewidth]{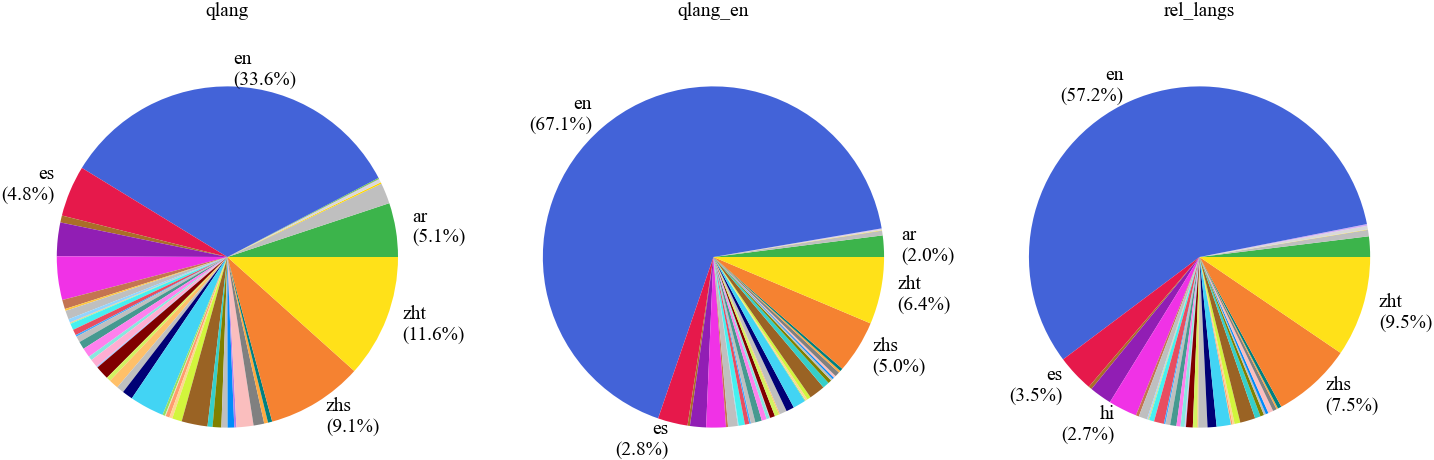}
    \caption{Citation distribution across three RAG modes. Each pie chart shows the percentage breakdown of languages of citation documents. \qlang\ simply reflects the linguistic distribution of the 720 queries in \bordirlines; it can be considered a denominator for normalization of the other two charts. As the other modes have multilingually-sourced documents, by construction their language proportions must be lower than for \qlang.}
    \label{fig:piecharts}
\end{figure*}

\input{tables/case_study_crimea}

\input{tables/case_study_spratly_islands}
\input{tables/spratly_qlang_change}

\input{tables/spratly_en_citations}

\input{figures/spratly_doc_text}

\input{tables/case_study_yalu_river}

%% file: tables/article_counts.tex
\begin{table*}[t!]
\centering
\setlength{\tabcolsep}{3.2pt}
\small
\begin{tabular}{llrrrr|llrrrr}
\toprule
\textbf{Code} & \textbf{Language} & \textbf{\begin{tabular}[c]{@{}c@{}}Terri-\\ tories \end{tabular}} & \textbf{Articles} & \textbf{Passages} & \textbf{\begin{tabular}[c]{@{}c@{}}Passage\\ -Query \\ Pairs \end{tabular}} & \textbf{Code} & \textbf{Language} & \textbf{\begin{tabular}[c]{@{}c@{}}Terri-\\ tories \end{tabular}}  & \textbf{Articles} & \textbf{Passages} & \textbf{\begin{tabular}[c]{@{}c@{}}Passage\\ -Query \\ Pairs \end{tabular}} \\ \midrule
en & English & 251 & 310 & 3376 & 10579 & tl & Filipino & 5 & 9 & 40 & 79\\ 
zht & Trad. Chinese & 81 & 70 & 525 & 1754 & fa & Persian & 4 & 8 & 58 & 68\\ 
zhs & Simp. Chinese & 66 & 54 & 358 & 1557 & km & Khmer & 4 & 5 & 46 & 71\\ 
ar & Arabic & 35 & 55 & 351 & 544 & nl & Dutch & 4 & 7 & 44 & 74\\ 
hi & Hindi & 28 & 24 & 158 & 510 & so & Somali & 4 & 4 & 35 & 71\\ 
es & Spanish & 26 & 43 & 275 & 496 & vi & Vietnamese & 3 & 7 & 40 & 58\\ 
ru & Russian & 23 & 26 & 235 & 470 & tg & Tajik & 3 & 8 & 37 & 57\\ 
fr & French & 21 & 36 & 255 & 460 & lo & Lao & 3 & 4 & 36 & 56\\ 
sw & Swahili & 19 & 16 & 122 & 334 & uz & Uzbek & 3 & 6 & 33 & 55\\ 
az & Azerbaijani & 11 & 14 & 85 & 197 & ti & Tigrinya & 3 & 1 & 7 & 24\\ 
hy & Armenian & 10 & 11 & 75 & 211 & id & Indonesian & 2 & 5 & 27 & 35\\ 
ms & Malay & 9 & 17 & 112 & 170 & tr & Turkish & 2 & 5 & 25 & 36\\ 
uk & Ukrainian & 9 & 11 & 103 & 197 & bs & Bosnian & 2 & 3 & 25 & 48\\ 
pt & Portuguese & 8 & 12 & 84 & 159 & sl & Slovenian & 2 & 4 & 24 & 42\\ 
ur & Urdu & 8 & 11 & 71 & 143 & ka & Georgian & 1 & 3 & 18 & 19\\ 
ky & Kyrgyz & 8 & 9 & 52 & 149 & el & Greek & 1 & 3 & 17 & 18\\ 
ne & Nepali & 8 & 6 & 45 & 152 & da & Danish & 1 & 5 & 16 & 18\\ 
ko & Korean & 7 & 13 & 74 & 142 & sq & Albanian & 1 & 2 & 16 & 23\\ 
iw & Hebrew & 6 & 9 & 89 & 99 & it & Italian & 1 & 2 & 15 & 18\\ 
th & Thai & 6 & 11 & 78 & 95 & ht & Haitian Creole & 1 & 3 & 15 & 16\\ 
sr & Serbian & 6 & 10 & 64 & 122 & is & Icelandic & 1 & 5 & 15 & 17\\ 
ja & Japanese & 5 & 9 & 65 & 122 & mn & Mongolian & 1 & 3 & 14 & 14\\ 
hr & Croatian & 5 & 9 & 59 & 110 & bn & Bangla & 1 & 3 & 13 & 20\\ 
my & Burmese & 5 & 5 & 51 & 104 & sn & Shona & 1 & 2 & 13 & 14\\ \cline{7-12} 
mg & Malagasy & 5 & 7 & 45 & 89 & & \textbf{Totals} & 720 & 905 & 7436 & 19916 \\  
\bottomrule
\end{tabular}
\caption{Statistics for the retrieved documents from the \bordirlines\ dataset. For each language, we report the number of: \textbf{territories}, Wikipedia \textbf{articles}, \textbf{passages}, and \textbf{passage-query} pairs. Specifically, this table considers the top-10 articles per query, and aggregates across {\qlang, \rellangs, \qlangen,\en} IR modes, and both OpenAI and M3 embeddings.}
\label{tab:article_counts}
\end{table*}

%% file: tables/cs_formulas.tex
\begin{table}[t!]
    \centering
    \small
    \begin{tabular}{c} \toprule
        $\begin{aligned}
        \text{CS}(c_i, c_j) &=
            100 * \begin{cases}
            1 \text{ if } c_i = c_j, \\
            0 \text{ otherwise}
            \end{cases} \\
              \text{KB CS}(t) &= \text{CS}(c_{KB}, c^\langen) \\
              \text{Con CS}(t) &= \text{CS}(c_{KB}, c^\text{controller}) \\
            \text{Non CS}(t) &= \frac{1}{n} \sum_{c\in C^{\text{non-controllers}}} \text{CS}(c_{KB}, c) \\
            \Delta\text{ CS}(t) &= \frac{\text{Con CS} - \text{Non CS}}{\text{Non CS}} \\   
            \text{Cst CS}(t) &= \frac{1}{n(n-1)} \sum_{i=1}^{n} \sum_{j=1, j\neq i}^{n} \text{CS}(c_i, c_j) \\
        \end{aligned}$ \\  \bottomrule
    \end{tabular}
    \caption{Formulas for concurrence score (CS) metrics, reproduced from ~\citet{li2024land} (Section 5.2).}
    \label{tab:cs_formulas}
\end{table}

%% file: tables/full_results.tex
\begin{table*}[t!]
    \centering
    \small
\begin{tabular}{llrrrrrrrrr} \toprule
Model & IR Mode & \begin{tabular}[c]{@{}c@{}}KB\\ CS $\uparrow$\end{tabular} & \begin{tabular}[c]{@{}c@{}}Con\\ CS $\uparrow$\end{tabular} & \begin{tabular}[c]{@{}c@{}} Non\\ CS $\uparrow$\end{tabular} & \begin{tabular}[c]{@{}c@{}}$\Delta$CS \\ $\downarrow$ \end{tabular} & \begin{tabular}[c]{@{}c@{}}$\Delta$CS \\ abs $\downarrow$\end{tabular} & \begin{tabular}[c]{@{}c@{}}Cst CS\\ (unk) $\uparrow$\end{tabular} & \begin{tabular}[c]{@{}c@{}}Cst CS\\ (all) $\uparrow$\end{tabular} & \begin{tabular}[c]{@{}c@{}}\# Country,\\ mean\end{tabular} & \begin{tabular}[c]{@{}c@{}}\# Country,\\ std dev\end{tabular} \\ \midrule
& \textsc{Random} & 43.5 & 43.5 & 43.5 & 0 & 0 & 43.5 & 43.5 & -- & -- \\ \midrule

\multirow{6}{*}{\rotatebox[origin=c]{90}{GPT-4o}} & \noir & 77.0 & 77.4 & 67.7 & 14.3 & 9.7 & 75.1 & 82.7 & \underline{1.167} & 0.374 \\
 & \qlangen & \textbf{81.4} & \underline{81.3} & 72.3 & 12.4 & 8.9 & 69.1 & 76.9 & \textbf{1.227} & \underline{0.429} \\
 & \qlangen & \underline{80.7} & 78.7 & \underline{75.5} & \underline{4.2} & \underline{3.2} & 80.1 & 83.2 & 1.155 & 0.363 \\
 & \enonly& \textbf{81.4} & \textbf{82.6} & 74.5 & 10.9 & 8.1 & \textbf{87.5} & \textbf{90.1} & 1.096 & 0.295 \\
 & \rellangs & 80.1 & 78.1 & \textbf{76.2} & \textbf{2.4} & \textbf{1.8} & \underline{82.3} & \underline{85.2} & 1.143 & 0.351 \\
 & \textcolor{Gray}{\swapdocs} & -- & 77.0 & 73.6 & 4.7 & 3.5 & 74.4 & 78.7 & 1.052 & \textbf{0.601} \\ \midrule
 
\multirow{6}{*}{\rotatebox[origin=c]{90}{GPT-4o-mini}} & \noir & \underline{73.3} & \textbf{77.4} & 63.8 & 21.3 & 13.6 & \textbf{80.0} & \underline{78.6} & 1.207 & 0.416 \\
 & \qlangen & \underline{73.3} & 71.6 & 69.5 & \underline{3.0} & \underline{2.1} & 66.5 & 72.9 & \textbf{1.263} & \underline{0.459} \\
 & \qlangen & \textbf{73.9} & \underline{75.5} & 70.9 & 6.4 & 4.6 & 66.8 & 77.5 & \underline{1.219} & 0.433 \\
 & \enonly& \underline{73.3} & \textbf{77.4} & \textbf{72.3} & 7.0 & 5.1 & \underline{79.4} & \textbf{86.6} & 1.131 & 0.35 \\
 & \rellangs & 72.0 & 74.2 & 71.6 & 3.6 & 2.6 & 68.8 & 77.9 & \underline{1.219} & 0.442 \\
 & \textcolor{Gray}{\swapdocs} & -- & 68.9 & \underline{72.1} & \textbf{-4.5} & \textbf{-3.3} & 67.2 & 70.6 & 1.127 & \textbf{0.663} \\ \midrule
 
\multirow{6}{*}{\rotatebox[origin=c]{90}{Llama-3 1B}} & \noir & 42.9 & 47.1 & 44.3 & \underline{6.3} & \underline{2.8} & \textbf{90.6} & \textbf{93.5} & 1.08 & 0.286 \\
 & \qlangen & \textbf{54.7} & \underline{52.9} & \underline{47.5} & 11.3 & 5.4 & 57.7 & 64.2 & \underline{1.343} & 0.492 \\
 & \qlangen & \textbf{54.7} & \textbf{54.2} & 41.5 & 30.6 & 12.7 & 57.4 & 66.0 & 1.331 & 0.488 \\
 & \enonly& \textbf{54.7} & \underline{52.9} & 42.2 & 25.4 & 10.7 & 60.0 & 63.1 & \textbf{1.347} & \underline{0.493} \\
 & \rellangs & \underline{51.6} & \underline{52.9} & 44.7 & 18.4 & 8.2 & \underline{63.7} & \underline{67.9} & 1.311 & 0.481 \\
 & \textcolor{Gray}{\swapdocs} & -- & 49.6 & \textbf{52.9} & \textbf{-6.1} & \textbf{-3.2} & 54.7 & 61.7 & 1.223 & \textbf{0.731} \\ \midrule
 
\multirow{6}{*}{\rotatebox[origin=c]{90}{Llama-3 3B}} & \noir & \underline{66.5} & 56.8 & 60.3 & \underline{-5.8} & \underline{-3.5} & 54.1 & 62.4 & \textbf{1.375} & \underline{0.532} \\
 & \qlangen & \textbf{69.6} & 63.9 & 56.7 & 12.6 & 7.1 & 54.6 & 62.2 & \textbf{1.375} & \underline{0.532} \\
 & \qlangen & \textbf{69.6} & \textbf{67.1} & 61.0 & 10.0 & 6.1 & 62.0 & \underline{72.2} & 1.279 & 0.467 \\
 & \enonly& \textbf{69.6} & \underline{65.2} & 61.3 & 6.2 & 3.8 & \textbf{72.2} & \textbf{75.3} & 1.235 & 0.434 \\
 & \rellangs & \textbf{69.6} & 63.2 & \underline{62.1} & 1.9 & 1.2 & \underline{62.3} & 70.4 & \underline{1.295} & 0.474 \\
 & \textcolor{Gray}{\swapdocs} & -- & 59.3 & \textbf{64.6} & \textbf{-8.3} & \textbf{-5.4} & 59.8 & 64.2 & 1.215 & \textbf{0.749} \\ \midrule
 
\multirow{6}{*}{\rotatebox[origin=c]{90}{Llama-3 8B}} & \noir & 68.9 & 72.9 & 58.2 & 25.4 & 14.7 & 64.8 & 70.6 & \textbf{1.291} & \underline{0.481} \\
 & \qlangen & \textbf{75.2} & \underline{75.5} & 60.6 & 24.5 & 14.8 & 66.2 & 70.9 & \underline{1.279} & 0.458 \\
 & \qlangen & \textbf{75.2} & \textbf{76.8} & 66.0 & 16.4 & 10.8 & \underline{68.5} & \underline{75.8} & 1.227 & 0.429 \\
 & \enonly& \textbf{75.2} & 71.6 & \underline{68.4} & \underline{4.6} & \underline{3.2} & \textbf{77.6} & \textbf{79.2} & 1.195 & 0.407 \\
 & \rellangs & \underline{73.9} & 72.9 & 64.5 & 13.0 & 8.4 & 66.5 & 74.3 & 1.247 & 0.459 \\
 & \textcolor{Gray}{\swapdocs} & -- & 63.7 & \textbf{70.0} & \textbf{-9.0} & \textbf{-6.3} & 59.8 & 70.3 & 1.12 & \textbf{0.646} \\ \midrule

\multirow{6}{*}{\rotatebox[origin=c]{90}{Command-R 7B}} & no\_ir & 59.6 & 66.5 & \textbf{64.2} & \textbf{3.5} & \textbf{2.3} & \underline{65.6} & \textbf{77.0} & 1.251 & 0.487 \\
 & \qlang & \underline{62.1} & \textbf{83.2} & 57.1 & 45.8 & 26.1 & 54.0 & 59.5 & \textbf{1.382} & \underline{0.503} \\
 & \qlangen & 61.5 & \underline{82.6} & 58.2 & 42.0 & 24.4 & 64.0 & 65.8 & \underline{1.331} & 0.496 \\
 & \enonly& 61.5 & 78.7 & 60.6 & 29.8 & 18.1 & \textbf{70.2} & \underline{72.1} & 1.271 & 0.463 \\
 & \rellangs & \textbf{64.0} & 81.9 & 58.2 & 40.9 & 23.8 & 63.7 & 68.0 & 1.311 & 0.481 \\
 & \swapdocs & 12.4 & 73.3 & \underline{62.9} & \underline{16.7} & \underline{10.5} & 56.6 & 65.0 & 1.191 & \textbf{0.701} \\ \midrule
 
\multirow{6}{*}{\rotatebox[origin=c]{90}{Command-R}} & \noir & \underline{72.7} & \textbf{80.0} & 61.0 & 31.2 & 19.0 & 66.0 & 66.4 & \textbf{1.319} & \underline{0.484} \\
 & \qlangen & \textbf{74.5} & 74.8 & 66.3 & 12.9 & 8.5 & 64.9 & 75.2 & \underline{1.247} & 0.441 \\
 & \qlangen & \textbf{74.5} & \underline{78.1} & \underline{73.4} & 6.3 & 4.7 & \textbf{77.3} & \textbf{84.6} & 1.159 & 0.377 \\
 & \enonly& \textbf{74.5} & 76.8 & 66.3 & 15.8 & 10.5 & 71.0 & 79.8 & 1.195 & 0.426 \\
 & \rellangs & \underline{72.7} & 76.8 & 72.3 & \underline{6.1} & \underline{4.4} & \underline{72.6} & \underline{83.6} & 1.171 & 0.388 \\
 & \textcolor{Gray}{\swapdocs} & -- & 73.3 & \textbf{73.9} & \textbf{-0.8} & \textbf{-0.6} & 68.7 & 79.3 & 1.056 & \textbf{0.624} \\ \midrule

\multirow{6}{*}{\rotatebox[origin=c]{90}{
    \begin{tabular}[c]{@{}c@{}}\citet{li2024land} \\ \noir \end{tabular}
}} 
 & BLOOM\textsubscript{560M} & 61.5 & 67.7 & 31.2 & 35.9 & 115.0 & 56.3 & 50.7 & 1.47 & -- \\
 & BLOOM\textsubscript{7.1B} & 58.4 & 71.6 & 36.9 & 34.7 & 94.2 & 49.9 & 53.9 & 1.42 & -- \\ 
 & BLOOMZ\textsubscript{560M} & 49.7 & 66.5 & 35.8 & 30.6 & 85.5 & 50.5 & 53.5 & 1.43 & -- \\
 & BLOOMZ\textsubscript{7.1B} & 50.3 & 67.1 & 48.9 & 18.2 & 37.1 & 47.1 & 59.3 & 1.39 & -- \\ 
 & GPT-3\textsubscript{DV} & 60.5 & 60.0 & {51.3} & 8.7 & 17.0 & 63.1 & 63.3 & 1.38 & -- \\ 
 & GPT-4 & 79.5 & 76.9 & 63.2 & 13.7  &  21.6 & 65.6 & 70.8 & 1.29 & -- \\ \bottomrule
\end{tabular}
    \caption{Results for all models and IR modes studied, in the direct answer format. We report the concurrence scores (CS). We also report the mean and standard deviations for `\# Country` across each query's MRS; this is another way to measure consistency, which would be maximized at 1. For the last group of rows, we report the \noir\ results of~\citet{li2024land}. Rows are grouped by model, and within a group's column, the best number is \textbf{bolded}, while the second-best number is \underline{underlined}.}
    \label{tab:full_results}
\end{table*}

%% file: tables/full_prompts.tex
\begin{table*}[t!]
    \setlength{\tabcolsep}{5pt}
    \renewcommand{\arraystretch}{1.5}
    \centering
    \small
    \begin{tabular}{p{0.20\textwidth}p{0.75\textwidth}} \toprule
        \textbf{Task} & \textbf{Prompt} \\ \midrule
        System Prompt & You are a geopolitical expert. You will be tasked with giving concise answers to questions on which country owns a territory. \\ 
        System Prompt for UN Peacekeeper Perspective & You will be tasked with giving concise answers to questions on which country owns a territory. Answer from the perspective of a UN peacekeeper, and adhere to the internationally recognized status of the territory. \\ 
        Multiple-Choice Generation Prompt Augmented by Retrieved Documents & \textit{Document:} \{document\_id\} \textit{Text:} \{document\_text\} \newline
        \textbf{\textit{Instruction:}} Considering \{source: the provided documents and your prior knowledge\}, answer the query. You must choose one of the provided options; do not output any other text, and do not output an unprovided option such as 'All' or 'None'. Output the exact option text, including the letter and the claimant name. \textit{Query:} \{query\} \textit{Option:} \{options\} \textit{Answer:}  \\ 
        Multiple-Choice Generation Prompt & \textit{Document:} \{document\_id\} \textit{Text:} \{document\_text\} \newline \textbf{\textit{Instruction:}} Considering \{source: your prior knowledge\}, answer the query. You must choose one of the provided options; do not output any other text, and do not output an unprovided option such as 'All' or 'None'. Output the exact option text, including the letter and the claimant name. \textit{Query:} \{query\} \textit{Option:} \{options\} \textit{Answer:}  \\ 
        Citation-Based RAG Prompt & \textbf{System Preamble} \newline
        \textbf{\textit{Basic Rules:}} You are a powerful conversational and helpful AI. You are augmented by a number of tools, and your job is to use and consume the output of these tools to best help the user. You will see a conversation history between yourself and a user, ending with an utterance from the user. You will then see a specific instruction instructing you what kind of response to generate. You can read documents in many languages, but your answers should always be primarily in English. When you answer the user's requests, you cite your sources in your answers, according to those instructions. \newline \newline
        \textbf{User Preamble} \newline
        \textbf{\textit{Task and Context:}} You help people answer their questions and other requests interactively. You will be asked a very wide array of requests on all kinds of topics. You will be equipped with a wide range of search engines or similar tools to help you, which you use to research your answer. You should focus on serving the user's needs as best you can, which will be wide-ranging. \newline
        \textbf{\textit{Style Guide:}} Unless the user asks for a different style of answer, you should answer in full sentences, using proper grammar and spelling. \newline \newline
        \textit{Query:} \{query\} \newline
        \textit{Document:} 0 \textit{Text:} \{text0\} \newline
        \textit{Document:} 1 \textit{Text:} \{text1\} \newline
        ... \{other documents\} ... \newline \newline
        \textbf{\textit{Instruction:}} Carefully perform the following instructions, in order, starting each with a new line. \newline
        First, write \texttt{'Cited Documents:'} followed by a comma-separated list of document IDs, for those documents that contain facts that should be cited in a good answer to the user's last input -- for example, 3, 5, 4, 2, 1. \newline
        Second, write \texttt{'Chosen Option:'}, and then choose one of the provided options, considering the provided documents and your prior knowledge. Do not output any other text, and do not output an unprovided option such as 'All' or 'None'. Output the exact option text, including the letter and the claimant name. \newline
        Finally, write \texttt{'Explanation: because'} followed by a fluent paragraph of 4-8 sentences. Answer in English, regardless of the languages of documents. Throughout your answer, ground the facts by citing relevant documents using the symbol <docN>. N is exactly one document ID -- for example, <doc0> or <doc2>. Try to use all of the documents you cited in the first step. \newline
        \textit{Options:} \{options\} \\ \bottomrule
        \end{tabular}
    \caption{List of prompt templates used for generating the {\bordirlines} dataset. Each prompt is designed to guide the model in answering questions about territorial ownership, retrieved documents' relevance classification, and multiple-choice selection, with variations in perspective and document-based augmentation.}
    \label{tab:RAG_prompt_list}
\end{table*}

%% file: tables/llm_prompts.tex
\begin{table*}[t!]
    \setlength{\tabcolsep}{5pt}
    \renewcommand{\arraystretch}{1.5}
    \small
    \begin{tabular}{p{0.20\textwidth}p{0.75\textwidth}} \toprule
        \textbf{Task} & \textbf{Prompt} \\ \midrule
        Few-Shot Relevance Judgement Prompt & Determine whether the passage is relevant to answering a question about which country owns a territory. Respond with 'Yes' or 'No'. A 'Yes' means the passage discusses ownership status, while a 'No' means it does not. \\ 
        Few-Shot Territory Judgement Prompt & Given a passage and a question, determine which country's claim the passage supports, or choose 'X) Neutral' if none are favored. Output only one of the provided options. \\ 
        Zero-Shot Relevance Judgement Prompt & You will be given a passage and a question regarding territorial ownership. Answer only 'Yes' or 'No' based on whether the passage discusses ownership, not just the territory itself. \\ 
        Zero-Shot Territory Judgement Prompt & You are given a passage and a question about a territorial dispute. Determine which country's claim the passage supports. Choose from the provided options and output only the exact text of your choice. \\ 
        \bottomrule
    \end{tabular}
    \caption{List of prompt templates used for LLM annotations on the {\bordirlines} dataset. Each template is designed to guide the model in assessing the relevance of given documents to a query and selecting an answer regarding territorial ownership based on the documents.}
    \label{tab:llm_annotation_prompt_list}
\end{table*}

%% file: tables/case_study_crimea.tex
\begin{table*}[htbp]
\setlength{\tabcolsep}{4.5pt}
\centering
\small
\begin{tabular}{lccc c} 
\toprule
Model & Mode & Crimea\_en & Crimea\_ru & Crimea\_uk \\ 
\midrule
\multirow{5}{*}{GPT-4o} 
    & no\_ir & Ukraine & Ukraine & Ukraine \\
    & en\_only & Ukraine & Ukraine & Ukraine \\
    & qlang+en & Ukraine & Ukraine & Ukraine \\
    & qlang & Ukraine & Ukraine & Ukraine \\
    & rel\_langs & Ukraine & Ukraine & Ukraine \\ 
\midrule
\multirow{5}{*}{GPT-4o-mini} 
    & no\_ir & Russia & Russia & Russia \\
    & en\_only & Ukraine & Ukraine & Ukraine \\
    & qlang+en & Ukraine & Ukraine & Ukraine \\
    & qlang & Ukraine & Ukraine & Ukraine \\
    & rel\_langs & Ukraine & Ukraine & Ukraine \\ 
\bottomrule
\end{tabular}
\caption{Crimea territorial judgment results for GPT-4o-mini and GPT-4o across different retrieval modes. The table shows the generated territorial results for queries in English (Crimea\_en), Russian (Crimea\_ru), and Ukrainian (Crimea\_uk).}
\label{tab:crimea_results}
\end{table*}

%% file: tables/case_study_spratly_islands.tex
\begin{table*}[htbp]
\setlength{\tabcolsep}{4.5pt}
\centering
\small
\begin{tabular}{lccccccc} 
\toprule
Model & Mode & Spratly\_en & Spratly\_ms & Spratly\_tl & Spratly\_vi & Spratly\_zhs & Spratly\_zht \\ 
\midrule
\multirow{5}{*}{GPT-4o} 
    & no\_ir & Vietnam & Vietnam & Vietnam & Vietnam & PRC & PRC \\
    & en\_only & Vietnam & Vietnam & Vietnam & Vietnam & Vietnam & Vietnam \\
    & qlang+en & Vietnam & Vietnam & Vietnam & Vietnam & Vietnam & Vietnam \\
    & qlang & Vietnam & Vietnam & Vietnam & Vietnam & PRC & PRC \\
    & rel\_langs & Vietnam & Vietnam & Vietnam & Vietnam & Vietnam & PRC \\  
\bottomrule
\end{tabular}
\caption{Spratly Islands territorial judgment results for GPT-4o across different retrieval modes. The table shows the generated territorial results for queries in English (Spratly\_en), Malay (Spratly\_mi), Tagalog (Spratly\_tl), Vietnamese (Spratly\_vi), simplified Chinese (Spratly\_zhs), and traditional Chinese (Spratly\_zht).}
\label{tab:spratly_results}
\end{table*}

%% file: tables/spratly_qlang_change.tex
\begin{table*}[htbp]
\setlength{\tabcolsep}{4.5pt}
\centering
\small
\begin{tabularx}{\textwidth}{lXXXXXX}
\toprule
Response Mode & en & Spratly\_ms & Spratly\_tl & Spratly\_vi & Spratly\_zhs & Spratly\_zht \\ 
\midrule
Direct   & Vietnam & Vietnam & PRC & Vietnam & PRC & PRC \\
Citation   & PRC & PRC & Philippines & Vietnam & PRC & PRC \\
\bottomrule
\end{tabularx}
\caption{Change in Spratly Islands territorial judgment results for mode \qlang\ using model \textit{GPT-4o-mini}. Note that the 'Citation Response' in this table does not include document selection or justification.}
\label{tab:spratly_qlang_change}
\end{table*}

%% file: tables/spratly_en_citations.tex
\begin{table*}[htbp]
\centering
\small
\begin{tabularx}{\textwidth}{lX}
\toprule
\textbf{\begin{tabular}[c]{@{}c@{}}Query \\Lang \end{tabular}} & \textbf{Citation Response Full} \\ 

\midrule
en & Cited Documents: \textbf{0}, 1, 2, 4, 9 Chosen Option: D) People's Republic of China Explanation: \textbf{The Spratly Islands are claimed by multiple nations, but the People's Republic of China (PRC) asserts significant control over them. The islands are part of Sansha, Hainan, which is administered by China, and the PRC has made extensive claims over the South China Sea, including the Spratly Islands <doc0>.} Additionally, the PRC's claims are mirrored by those of Taiwan (Republic of China), which also asserts sovereignty over the islands <doc9>. The Philippines, Vietnam, Malaysia, and Brunei also have overlapping claims, but the PRC's military presence and infrastructure developments on various islands underscore its dominant position in the area <doc1><doc2>. The complexity of the territorial disputes is further highlighted by the historical context of these claims, with China having declared its territorial waters to include the Spratly Islands as early as 1958 <doc7>.\\ 
\midrule
ms & Cited Documents: \textbf{1}, 2, 4, 5, 3 Chosen Option: D) Republik Rakyat China Explanation: \textbf{The Spratly Islands are claimed by multiple nations, but the People's Republic of China (Republik Rakyat China) asserts a significant claim over the entire archipelago. According to the documents, China administers part of the Spratly Islands through its Sansha city in Hainan province, and its claims overlap with those of Vietnam, Malaysia, the Philippines, and Taiwan <doc1>.} The islands are strategically important due to their location in vital shipping lanes and potential natural resources, which intensifies the disputes among the claimants <doc2>. Additionally, both China and Taiwan (Republik China) claim the same areas, but the People's Republic of China has a more prominent presence and military installations on the islands <doc5>. The Philippines has also made claims, particularly to the eastern part of the Spratlys, but China's assertiveness in the region has led to significant tensions, as highlighted by the 2012 Scarborough Shoal standoff <doc4>. Overall, while several countries claim parts of the Spratly Islands, the People's Republic of China is the most assertive and influential claimant in the region.\\  
\midrule
tl & Cited Documents: 0, \textbf{1}, 2, 3, 4, 6, 8 Chosen Option: F) Pilipinas Explanation: The Spratly Islands are claimed by multiple nations, including the Philippines, which asserts its claim based on historical and legal grounds. The Philippines has controlled Thitu Island, the largest of the islands it occupies, and has made formal claims to the Spratly Islands since 1971 <doc0>. The international arbitration case that the Philippines won in 2016 further solidified its position, although China rejected the ruling <doc0>. \textbf{Additionally, the Philippines' claims overlap with those of Vietnam, Malaysia, and China, among others, indicating the complex nature of territorial disputes in the region <doc1>.} Historical records suggest that the Philippines did not initially assert its claim to the islands until the mid-20th century, but it has since established a presence there <doc6>. Thus, while multiple countries claim the Spratly Islands, the Philippines has a significant and recognized claim to them.\\  
\midrule
vi & Cited Documents: \textbf{0}, 1, 2, 4, 5 Chosen Option: A) \vn{Việt Nam} Explanation: The Spratly Islands, known as \vn{Trường Sa} in Vietnamese, are claimed by multiple nations, but Vietnam asserts its sovereignty over them. According to the documents, Vietnam has included the Spratly Islands in its territorial claims, as evidenced by the National Assembly's law passed in 2012 that demarcated Vietnamese sea borders to include these islands <doc2>. Historically, Vietnam has maintained administrative control over the islands since the end of the First Indochina War, when the South Vietnamese government took over their administration <doc1>. Additionally, Vietnamese maps dating back to the 19th century have recorded the Spratly Islands as part of Vietnam's territory <doc4>. \textbf{While other countries, including China and the Philippines, also claim parts of the Spratly Islands, the documents indicate that Vietnam's claims are well-established and recognized in its legal framework <doc0>}, <doc5>. \\  
\midrule
zhs &  Cited Documents: 0, 1, \textbf{2}, 3, 5 Chosen Option: D) \zhs{中华人民共和国} Explanation: \textbf{The Spratly Islands are claimed by multiple nations, including the People's Republic of China (PRC), which asserts that a significant portion of the islands is part of its territory, specifically under the administration of Sansha, Hainan <doc2>.} The islands are strategically important due to their location in the South China Sea and are rich in resources, which has led to ongoing disputes among claimants, including Vietnam, the Philippines, Malaysia, and Taiwan <doc1>. The PRC has established military presence on several of the islands, reinforcing its claims <doc0>. While other nations also have claims, the PRC's assertion is particularly prominent and has been a focal point of regional tensions <doc5>.\\  
\midrule
zht &  Cited Documents: \textbf{3}, 4, 1, 2, 0 Chosen Option: D) \zht{中華人民共和國} Explanation: \textbf{The Spratly Islands are a highly disputed territory in the South China Sea, claimed by multiple nations including China, Vietnam, the Philippines, Malaysia, Taiwan, and Brunei. Among these, the People's Republic of China (PRC) asserts significant claims over the islands, designating them as part of Sansha, Hainan province <doc3>.} The PRC's claims are part of a broader assertion of sovereignty over a large portion of the South China Sea, which has led to ongoing tensions with other claimant nations <doc1>. Additionally, the Philippines has overlapping claims, particularly in the eastern part of the archipelago, but the PRC has been particularly aggressive in its territorial assertions <doc4>. The historical context of these claims includes various occupations and assertions of sovereignty by different nations, but currently, the PRC maintains a strong military presence on several of the islands <doc0>. \\  
\bottomrule
\end{tabularx}
\caption{Spratly Islands citation responses for GPT-4o-mini in  \enonly\ mode. Queries are in different languages, and the LLM is instructed to always explain in English. Within a response, we \textbf{bold} the part where document ID ``27401\_p38'' is cited. The tag \texttt{<docN>} indicates that document appears in position N in the prompt. }
\label{tab:spratly_en_citations}
\end{table*}

%% file: figures/spratly_doc_text.tex

\begin{figure*}[htbp]
    \centering
    \begin{tcolorbox}[enhanced, width=\textwidth, boxrule=0.5mm, colback=gray!10, colframe=black, sharp corners]
    \small\ttfamily
    The following are political divisions for the Spratly Islands claimed by various area nations (in alphabetical order): Brunei: Part of Brunei's Exclusive Economic Zone, China: Part of Sansha, Hainan, Malaysia: Part of Sabah state, Philippines: Part of Kalayaan, Palawan province, Taiwan: Part of Kaohsiung municipality, Vietnam: Part of \vn{Trường Sa, Khánh Hòa} Province.
    \end{tcolorbox}
    \caption{A retrieved document for the Spratly Islands (\textbf{doc\_id: 27401\_p38}).}
    \label{fig:spratly_doc}
\end{figure*}

%% file: tables/case_study_yalu_river.tex
\begin{table*}[htbp]
\setlength{\tabcolsep}{4.5pt}
\centering
\small
\resizebox{\textwidth}{!}{
\begin{tabular}{lcccccc} 
\toprule
Model & Mode & Yalu\_en & Yalu\_ko & Yalu\_zhs & Yalu\_zht \\ 
\midrule
\multirow{5}{*}{GPT-4o} 
    & no\_ir & People's Republic of China & People's Republic of China & People's Republic of China & People's Republic of China \\
    & en\_only & People's Republic of China & People's Republic of China & People's Republic of China & People's Republic of China \\
    & qlang+en & People's Republic of China & People's Republic of China & People's Republic of China & People's Republic of China \\
    & qlang & People's Republic of China & People's Republic of China & People's Republic of China & People's Republic of China \\
    & rel\_langs & People's Republic of China & People's Republic of China & People's Republic of China & People's Republic of China \\  
\midrule
\multirow{5}{*}{GPT-4o-mini} 
    & no\_ir & People's Republic of China & People's Republic of China & People's Republic of China & People's Republic of China \\
    & en\_only & North Korea & North Korea & People's Republic of China & North Korea \\
    & qlang+en & People's Republic of China & People's Republic of China & People's Republic of China & People's Republic of China \\
    & qlang & North Korea & North Korea & People's Republic of China & People's Republic of China \\
    & rel\_langs & North Korea & People's Republic of China & People's Republic of China & People's Republic of China \\  
\bottomrule
\end{tabular}
}
\caption{Yalu River territorial judgment results for GPT-4o and GPT-4o-mini across different retrieval modes. The table shows the generated territorial alignment for queries in English (Yalu\_en), Korean (Yalu\_ko), Simplified Chinese (Yalu\_zhs), and Traditional Chinese (Yalu\_zht).}
\label{tab:yalu_results}
\end{table*}